\documentclass[letterpaper,twocolumn,10pt]{article}
\usepackage{usenix2019_v3}
\usepackage{tikz}
\usepackage{amsmath}

\usepackage{subfig}
\usepackage[indLines=false]{algpseudocodex}
\usepackage{algorithm}
\usepackage{xcolor}
\usepackage{cleveref}
\usepackage[textsize=tiny]{todonotes}

\usepackage{pifont}
\usepackage{amsmath}
\usepackage{xcolor}
\usepackage{colortbl}  
\usepackage{multirow}
\usepackage{booktabs}
\usepackage{amsfonts}
\usepackage{nicefrac}
\usepackage{adjustbox}
\usepackage{mdframed}
\usepackage{amssymb}

\usepackage{xcolor} 
\usepackage[normalem]{ulem} 

\usepackage[textsize=tiny]{todonotes}
\DeclareMathOperator*{\argmin}{arg\,min}

\newlength\myindent
\NewDocumentCommand{\statcirc}{ O{#2} m }{%
    \begin{tikzpicture}
    \node[circle,minimum width=2mm,draw,fill=#1] {};
    \fill[#2] (0,0) circle (1.0ex); 
    \fill[#1] (0,0) -- (90:1ex) arc (90:270:1ex) -- cycle; 
    \end{tikzpicture}
}

\algblockdefx[game]{game}{EndGame}[2]{\textbf{experiment} \textsc{#1}(#2)}{\algpx@endIndent}
\algtext*{EndGame}

\let\oldReturn\Return
\renewcommand{\Return}{\State\oldReturn}

\newif\ifanonymous
\begin{document}

\date{}

\title{PTW: Pivotal Tuning Watermarking for Pre-Trained Image Generators}


\ifanonymous
    \author{\rm Anonymous Author(s)*}
\else
    \author{
    {\rm Nils Lukas}\\
    University of Waterloo
    \and
    {\rm Florian Kerschbaum}\\
    University of Waterloo
    }
\fi

\maketitle

\begin{abstract}
Deepfakes refer to content synthesized using deep generators, which, when \emph{misused}, have the potential to erode trust in digital media.
Synthesizing high-quality deepfakes requires access to large and complex generators only a few entities can train and provide.
The threat is malicious users that exploit access to the provided model and generate harmful deepfakes without risking detection.
Watermarking makes deepfakes detectable by embedding an identifiable code into the generator that is later extractable from its generated images. 
We propose Pivotal Tuning Watermarking (PTW), a method for watermarking pre-trained generators (i) three orders of magnitude faster than watermarking from scratch and (ii) without the need for any training data. 
We improve existing watermarking methods and scale to generators $4 \times$ larger than related work. 
PTW can embed longer codes than existing methods while better preserving the generator's image quality. 
We propose rigorous, game-based definitions for robustness and undetectability and our study reveals that watermarking is not robust against an adaptive white-box attacker who has control over the generator's parameters. 
We propose an adaptive attack that can successfully remove any watermarking with access to only $200$ non-watermarked images.
Our work challenges the trustworthiness of watermarking for deepfake detection when the parameters of a generator are available.\footnote{This work has been accepted to USENIX Security 2023.}
%
\end{abstract}

\begin{figure*}
    \centering
    \includegraphics[width=1.\linewidth]{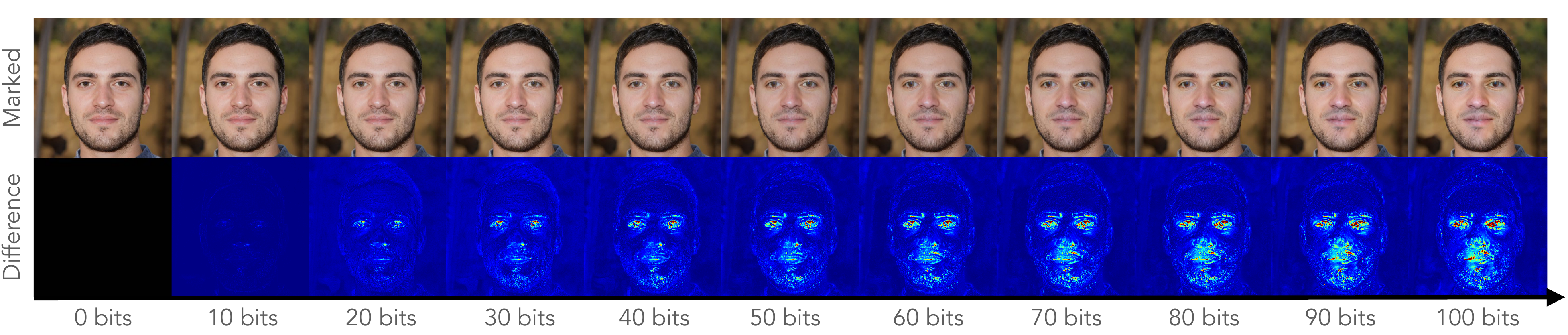}
    \caption{A demonstration of our watermark and the impact of the number of embedded bits on the visual image quality. 
    The top row shows the watermarked, synthetic image and the bottom row shows its difference to the same image without a watermark. }
    \label{fig:teaser}
\end{figure*}

\section{Introduction}
Deepfakes, a term that describes synthetic media generated using deep image generators, have received widespread attention in recent years. 
While deepfakes offer many beneficial use cases, for example, in scientific research~\cite{crystal2020photographic,sorin2020creating} or education~\cite{prezja2022deepfake, silbey2018upside,engelhardt2019cross}, they have also raised ethical concerns because of their potential to be \emph{misused} which can lead to an erosion of trust in digital media. 
Deepfakes have been scrutinized for their use in disinformation campaigns~\cite{agarwal2019protecting, hine2022new}, impersonation attacks~\cite{mink2022deepphish,dong2022protecting} or when used to create non-consensual media of an individual violating their privacy~\cite{harwell2018scarlett, de2021distinct}.
These threats highlight the need to control the misuse of deepfakes.

While some deepfakes can be recreated using traditional computer graphics, deep learning such as the Generative Adversarial Network (GAN)~\cite{goodfellow2014generative} can reduce the time and effort needed to create deepfakes.
However, training GANs require a significant investment in terms of computational resources~\cite{karras2021alias} and data preparation, including collection, organization, and cleaning.
These costs make training image generators a prohibitive endeavor for many.
Consequently, generators are often trained by one \emph{provider} and made available to many users through Machine-Learning-as-a-Service~\cite{cai2021generative}.
The provider wants to disclose their model responsibly and deter \emph{model misuse}, which is the unethical use of their model to generate harmful or misleading content~\cite{mirsky2021creation}.

 \textbf{Problem. } Consider a provider who wants to make their image generator publicly accessible under a contractual usage agreement that serves to prevent misuse of the model.
The threat is a user who breaks this agreement and uses the generator to synthesize and distribute harmful deepfakes without detection.
To mitigate this threat in practice, companies like OpenAI have deployed invasive prevention measures by providing only monitored access to their models through a black-box API. 
Users that synthesize deepfakes are detectable when they break the usage agreement if the provider matches the deepfake with their database.
This helps deter misuse of the model, but it can also lead to a lack of transparency and limit researchers and individuals from using their technology~\cite{tamkin2021understanding,ding2023towards}. 
For example, query monitoring, which is used in practice by companies such as OpenAI, raises privacy concerns as it involves collecting and potentially storing sensitive information about the user's queries.
A better solution would be to implement methods that deter model misuse without the need for query monitoring. 

A potential solution is to rely on deepfake detection methods~\cite{li2020face, jeong2022bihpf, chen2023watching}. 
The idea guiding such methods is to exploit artifacts from synthetic images that separate fake and real content. 
While these detectors protect well against some deepfakes, it has been demonstrated that they can be bypassed by unseen, improved generators that adapt to existing detectors~\cite{dong2022think}. 
As technology advances, it is possible that generators will be developed that synthesize virtually indistinguishable images, rendering passive deepfake detection methods ineffective in the long term. 

A different approach to deepfake detection is watermarking~\cite{yu2021artificial} when the detection method can access and modify the \emph{target generator}. 
This is a kind of watermarking that modifies the generator to embed an identifiable message that is later extractable from access to the generated content using a secret key. 
Methods such as generator watermarking~\cite{yu2019attributing, yu2021artificial} remain applicable to unseen image generators that could be developed in the future. 
For deepfake detection, the provider needs a watermark that is extractable from any image synthesized by the generator.
We refer to this setting as a \emph{no-box} verification because the verifier only requires access to the generated content but not to the generator model. 
However, there are still several challenges in designing no-box watermarks. 
These include (i) the ability to embed long messages with limited impact on the model's utility, (ii) the undetectability of the watermark without the secret key, (iii) robustness against removal, and (iv) the method should be efficient.

\textbf{Solution Overview.} 
Existing watermarking methods are difficult to scale to high-resolution models because they require re-training the generator from scratch, which is computationally intensive~\cite{yu2019attributing, yu2021artificial}.
Moreover, while some existing watermarks claim a good capacity/utility trade-off~\cite{yu2021artificial}, these claims have been limited to relatively small generators. 
To address these challenges, we propose an efficient watermark embedding called Pivotal Tuning Watermarking (PTW).
PTW is the first method to embed watermarks into pre-trained generators and speeds up the embedding process by three orders of magnitude from more than one GPU month when watermarking from scratch~\cite{yu2019attributing, yu2021artificial} to less than one hour.  
We identify modifications to existing watermarks that speed up their embedding significantly and propose our own watermark that can be embedded using PTW with an improved capacity/utility trade-off compared to related work. 
\Cref{fig:teaser} visualizes this trade-off for our watermark. 

We propose game-based definitions for robustness and undetectability and evaluate our watermark in two threat models, where the adversary either has access to the generator only through an API (\emph{black-box}) or has control over its parameters (\emph{white-box}).
Our results confirm that existing watermarking methods~\cite{yu2021artificial,yu2019attributing} are robust and undetectable in the black-box threat model using existing attacks. 
We propose three new attacks: a black-box attack called Super-Resolution and two white-box attacks called (1) Overwriting and (2) Reverse Pivotal Tuning.
Our experiments indicate that watermarking can be robust in the black-box setting when the attacker has limited access to a high-quality image generator. 
However, it cannot withstand a white-box attacker with access to only $200$ images ($\approx0.3\%$ of the generator's training dataset) who can remove watermarks at a negligible loss in the generator's image quality.

\subsection{Contributions}
\begin{itemize} \itemsep0mm
    \item We propose a watermarking method for pre-trained generators called Pivotal Tuning Watermarking (PTW). 
    PTW does not require any training data, and (ii) is three orders of magnitude computationally less intensive than existing methods~\cite{yu2019attributing,yu2021artificial}
    \item We modify existing watermarking methods~\cite{yu2021artificial,yu2019attributing} for GANs to be compatible with pre-trained generators, enabling their replication efficiently for large models.
    \item We provide game-based definitions for robustness and undetectability.
    \item We propose experiment with three generator architectures (StyleGAN2~\cite{karras2019style}, StyleGAN3~\cite{karras2021alias} and StyleGAN-XL~\cite{sauer2022stylegan}) on multiple high-quality image generation datasets.
    \item We propose black-box and white-box watermark removal attacks.
    Our results show that watermarking is not robust in practice against our white-box attacks. 
    \item We make our source code publicly available that implements all presented experiments\footnote{\url{https://github.com/nilslukas/gan-watermark}}.
\end{itemize}

\section{Background \& Related Work}
\label{sec:background-and-related-work}
This section provides a background on generative models and Pivotal Tuning~\cite{roich2022pivotal}, followed by a description of related work on the detection and attribution of deepfakes.
\subsection{Background}
\label{sec:background}
\textbf{Generative Adversarial Network (GAN).}
GANs~\cite{goodfellow2014generative}  define a generator $G: \mathcal{Z} \rightarrow \mathcal{X}$ that maps from a latent space $\mathcal{Z}$ to images $\mathcal{X}$ and a discriminator $F: \mathcal{X} \rightarrow \{0,1\}$ that maps images to binary labels.
The labels represent \emph{real} and \emph{fake} images.
Let $D \subseteq \mathcal{X}$ be an image dataset and let $\theta_{\text{F}},\theta_{\text{G}}$ be parameters for a discriminator and generator.
The unsaturated logistic loss for GANs is written as follows.
\begin{align}
    \label{eq:gan_loss}
     \mathcal{L}_{GAN} =& \underset{x \sim D}{\mathbb{E}}[\log F(\theta_F, x)]  \\ &+ \underset{z \sim \mathcal{N}(0,I^d)}{\mathbb{E}}[\log(1-F(\theta_F, G(\theta_G,z))]\nonumber
\end{align}
During training, the discriminator learns to classify real and fake images, and the generator learns to fool the discriminator.

\textbf{StyleGAN}~\cite{karras2019style}. 
The StyleGAN is a specific GAN architecture that introduces a mapper $f: \mathcal{Z} \rightarrow \mathcal{W}$, which maps latent codes into an intermediate latent space. 
This intermediate latent space contains \emph{styles} with fine-grained control over the synthesized image. 
Since its inception, the basic StyleGAN~\cite{karras2019style} has been revised many times leading to the development of StyleGAN2~\cite{karras2020analyzing}, StyleGAN3~\cite{karras2021alias} and recently StyleGAN-XL~\cite{sauer2022stylegan}.
These generators achieve state-of-the-art performance on many image generation datasets, including ImageNet~\cite{deng2009imagenet} where they outperform\footnote{\url{https://paperswithcode.com/dataset/ffhq}} other publicly accessible generators such as Latent Diffusion models~\cite{rombach2022high}.

\textbf{Pivotal Tuning}~\cite{roich2022pivotal}. Pivotal Tuning is a known method to regularize a pre-trained generator while preserving a high fidelity to the generator before tuning.
The idea is to preserve the mapping from latent codes to images by cloning and freezing the generator's parameters, referred to as the \emph{Pivot} with parameters $\theta_G$, and then fine-tuning a trainable, second generator $\theta_G^*$ with some regularization term $R(\cdot)$. 
It has been demonstrated that Pivotal Tuning achieves near-perfect image inversion while enabling latent-based image editing~\cite{roich2022pivotal}.
The Pivotal Tuning loss $\mathcal{L}_{\text{PT}}$ is written as follows.
\begin{align}   
    \label{eq:pivotal_tuning}
    \mathcal{L}_{\text{PT}} = \mathcal{L}_{\text{LPIPS}}(x_0, x) + \lambda_{\text{R}} R(x)
\end{align}
where $x_0=G(z, \theta_0)$ is an image synthesized by the frozen Pivotal generator using a latent code $z\in \mathcal{Z}$ and $x=G(z,\theta_1)$ is the image generated with the cloned weights $\theta_1$ for the same latent code. 
The Learned Perceptual Image Patch Similarity (LPIPS)~\cite{zhang2018unreasonable} loss $\mathcal{L}_{\text{LPIPS}}$ quantifies a perceptual similarity between images extracted using deep feature extractors. 
%
\subsection{Related Work}
%
This section summarizes related work on deepfake detection and attribution for deep image generators.

\textbf{Deepfake Detection and Attribution.} 
Deepfake detection and attribution are the tasks of identifying fake images generated or manipulated using deep image generators. 
Detection focuses only on detecting whether an image is fake, whereas attribution focuses on determining the image's origin. 
For a given \emph{target generator} that is used to synthesize a deepfake, we taxonomize 
existing work by (i) the level of access to this target generator and (ii) whether the target generator's parameters can be modified by the detection or attribution method before the deployment of the generator. 

\textbf{(i) Without Generator Access}: In this setting, the detection algorithm does not have access to the target generator.
Existing work on detecting deepfakes trains classifiers on a public set of deepfakes with known labels for fake/real images~\cite{rossler2019faceforensics++, li2020face, dolhansky2020deepfake}, exploit semantic incoherence such as asymmetries~\cite{matern2019exploiting, hu2021exposing} or low-level artifacts from the generation process~\cite{qian2020thinking, jeong2022bihpf, marra2019gans, chen2023watching,frank2020leveraging,wu2020sstnet}. 
Deepfake attribution methods without access to the target generator apply unsupervised learning methods~\cite{yu2019attributing, girish2021towards} or only attribute deepfakes to an architecture (and not a generator instance)~\cite{yang2022deepfake}.
Although these methods have proven effective in detecting some deepfakes, it has been shown that they can be evaded by an adversary who adapts to these detectors~\cite{dong2022think}.

\textbf{(ii) With Generator Access}: The detection method can have some level of access to the generator, including black-box API or white-box access to its parameters. 
\textbf{(Fingerprinting)} Methods that do not modify the generator's parameters are referred to as \emph{fingerprinting} methods. 
Recently, attribution methods have been proposed for Latent Diffusion models~\cite{sha2022fake} based on training classifiers on the model's generated data. 
For GANs, fingerprinting methods rely on training classifiers on the target generator's data~\cite{bui2022repmix,yang2021learning,yu2019attributing}.
\textbf{(Watermarking)} Methods that modify the generator before deployment are referred to as \emph{watermarking} methods.
Yu et al.~\cite{yu2021artificial} modify the generator's training data and re-train a watermarked generator. 
Another approach is to modify the generator's training procedure~\cite{yu2020responsible}. 
All existing methods require training the generator from scratch to embed a watermark. 

\section{Threat Model}
\label{sec:threat_model}
\begin{figure}
    \centering
    \includegraphics[width=1\linewidth]{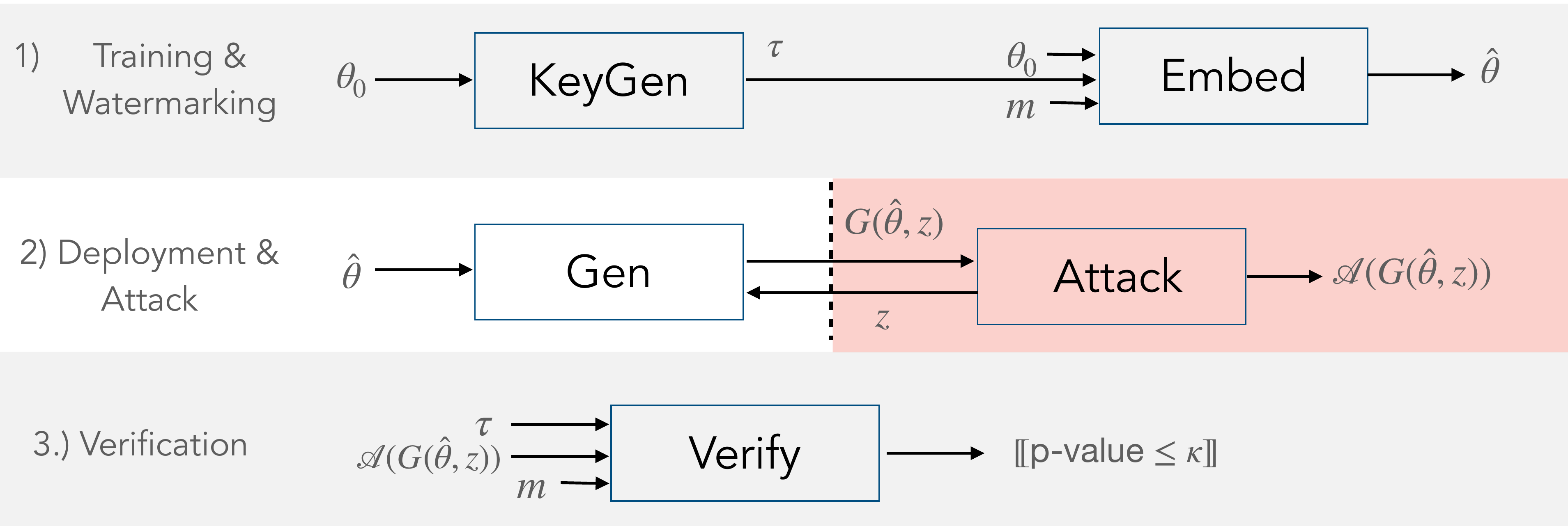}
    \caption{Deepfake detection in a \emph{no-box} setting by watermarking the generator. }
    \label{fig:ptw-overview}
\end{figure}
Our threat model consists of a defender, who we also refer to as the model provider, and an adversary, who controls a malicious user. 
We define the following auxiliary functions.
\begin{itemize}
    \itemsep0mm
    \item $x \leftarrow \textsc{Gen}(\theta_0; z)$: A function to generate\footnote{We write $\textsc{Gen}(\theta_0)$ to denote an image generated with $\theta_0$ on a randomly sampled latent code $z \sim \mathcal{Z}$.} an image on a latent code $z$ from a generator $\theta_0$. 
    \item $\tau \leftarrow \textsc{KeyGen}(\theta_0)$: A randomized function to generate a watermarking key given a generator.
    \item $\hat{\theta} \leftarrow \textsc{Embed}(\theta_0, \tau, m)$: Given a generator $\theta_G$, a secret watermarking key $\tau$ and a message $m\in \mathcal{M}$, this function returns parameters $\hat{\theta}$ of a watermarked generator.
    \item $p \gets \textsc{Verify}(x, \tau, m)$: Extracts a message from an image $x$ and calculates a $p$-value to reject the null hypothesis that both messages match by chance. 
\end{itemize}
Using the functions defined above, \Cref{fig:ptw-overview} illustrates attribution through watermarking in three steps.
First, the provider trains a generator and embeds a watermark. 
Second, the generator is deployed, and a malicious user generates harmful deepfakes at any time after deployment. 
Finally, deepfakes are attributed to the generator by verifying their watermark.

\textbf{Adversary's Capabilities.} We consider two adversaries that differ in their level of access to the target generator. 
Our first adversary, the \emph{black-box} adversary, has only API access to the target generator.
This means they can query the generator on any latent code $z\in \mathcal{Z}$, but do not have knowledge of its parameters or intermediate activations. 
The black-box adversary is limited in the number of queries to the generator since queries usually incur a monetary cost to the user. 
Our second adversary, the \emph{white-box} adversary, has full access to the (watermarked) target generator's parameters, meaning they can tune the generator's weights in an attempt to remove watermarks and generate any number of watermarked images. 
We use the same definitions for adversaries with black-box API access and white-box access as Lukas et al.~\cite{lukas2022sok}.

Both adversaries can access a limited set of $R$ real, non-watermarked images from the same distribution as the defender's training data. 
An adversary can have limited access to real images without a watermark, which is a commonly made assumption to evaluate the robustness of watermarking~\cite{lukas2022sok}.
We assume limited availability of real, non-watermarked data in our threat model, as an attacker with sufficient data and computational resources could train and deploy their own generator without a watermark. 
A black-box attacker could attempt to \emph{extract} the generator~\mbox{\cite{szyller2021good}} by training a surrogate on generated data, but this is out of the scope of our work since model extraction (i) requires massive computational resources and (ii) likely results in surrogates with low quality~\cite{lukas2022sok}.

\begin{table}[htpb]
    \small
    \centering
    \begin{tabular}{@{}p{80pt}@{~~}l@{}}
    \toprule
    \bf Notation                         & \bf Description \\
    \midrule
    $\mathcal{M}$                        & Distribution over messages\\
    
    $\mathcal{D}$                        & Distribution over images \\
    $\mathcal{D}^n$                      & Distribution over $n$ images \\
    $D \sim \mathcal{D}$                 & Draw images $D$ uniformly from $\mathcal{D}$ \\
    $y \gets \mathcal{P}(\vec{x})$       & Call $\mathcal{P}$ with $\vec{x}$ and assign result to $y$ \\
    $\mathcal{A}$                        & A procedure denoting an adversary \\
    $\theta$                               & Parameters of a generator \\
    $m$                               & A watermarking message \\
    $\tau$                               & A secret watermarking key\\
    
    \bottomrule
    \end{tabular}
    \caption{A summary of this chapter's notation.\label{tab:notation}}
\end{table}

\textbf{Adversary's Objective.} The common goal of our adversaries is to synthesize images (i) with high visual fidelity to real images and (ii) to generate images that do not retain the watermark. 
An image does not retain its watermark if the verification mistakenly outputs zero. 

\textbf{Defender's Capabilities.} The defender has access to their own generator's parameters and the secret watermarking key. 
Their objective is to verify any given image whether it originated from their generator.
We refer to the defender's access during verification as \emph{no-box} because, unlike black-box watermarking~\cite{adi2018turning} that can verify the watermark using many queries to the model, with no-box access, the watermark needs to be verified using only one generated image and without control over the query used to generate the image.

\subsection{Robustness}

\Cref{alg:watermark-verification-game} encodes the watermark robustness game given a pre-trained generator $\theta_0$, a confidence threshold for the verification $\kappa\in [0,1]$, a number $R$ of non-watermarked images available to the attacker and a challenge size $K\in \mathbb{N}^+$. 
We choose $\kappa = 0.05$. 
The challenge size is the number of images that the adversary has to synthesize to win the game. 
Note that an adversary with auxiliary access to at least $K$ non-watermarked, real images can always trivially win our security game by returning these real images in their attack.  
For this reason, we make the assumption that the adversary's auxiliary dataset size $R \ll K$ is much smaller than the challenge size. 
We choose K=$50\,000$, which allows comparing the quality of the generated images with related work~\mbox{\cite{karras2020analyzing, karras2021alias}}.

\begin{algorithm}
\caption{Watermark Robustness Game}
\small
\begin{algorithmic}[1]
\Procedure{Access}{$\theta$} \Comment{Determines adversary's access}
    \Return $\theta$ if white-box else $G(\cdot; \theta)$ \Comment{Returns full model or black-box generator}
\EndProcedure
\end{algorithmic}
\begin{algorithmic}[1]
\game{Robustness}{$\theta_0, \kappa, R, K$}     
    \State $\tau \gets \textsc{KeyGen}(\theta_0)$ \Comment{Generate watermark key}
    \State $m \sim \mathcal{M}$ \Comment{Sample watermark message}
    \State $\theta_1\gets \textsc{Embed}(\theta_0, \tau, m)$ \Comment{Embed watermark into model}   
    \State $D_{\mathcal{A}} \sim \mathcal{D}^R$ \Comment{Attacker's dataset of R images}
    \State $X^0, X^1, Y \gets \varnothing$ \Comment{Initialize sets}
    \State $B \sim \{0,1\}^K$ \Comment{K random coin flips} 
    \For{$i \in \{1..K\}$}
        \State $X^0 \gets X^0 \cup \{\mathcal{A}(\textsc{Access}(\theta_1), D_{\mathcal{A}})\}$ \Comment{Evasion}
        \State $X^1 \gets X^1 \cup \{\textsc{Gen}(\theta_0,z) | z\sim \mathcal{Z}\}$ \Comment{Without watermark}
        \State $Y \gets Y \cup \left\{\begin{array}{ll}
        B_i & \text{if } \textsc{Verify}(X^{B_i}_i, \tau, m) \leq \kappa \\
        1-B_i & \text{otherwise}
        \end{array}\right\}$
    \EndFor
    \State $e \gets  \frac{1}{K} \sum_{i \in \{1..K\}} (1-Y_i)$ \Comment{Compute evasion rate}

    \Return $X^0, e$ \Comment{Return images after evasion and evasion rate}
\EndGame
\end{algorithmic}
\label{alg:watermark-verification-game}
\end{algorithm}

In \Cref{alg:watermark-verification-game}, the game generates a watermark key $\tau$ and a watermarking message $m\in \mathcal{M}$ uniformly at random. (lines 2-3).
The defender then embeds the watermark into the generator $\theta_0$ to obtain the watermarked generator $\theta_1$ (line 4).
The adversary obtains access to $R$ non-watermarked images (line 5), and $K$ unbiased coin flips dictate the sequence of whether the verifier gets access to the watermarked image after evasion or the non-watermarked image (line 7). 
A correct verification means that the $p$-value returned by the verification function is at most $\kappa$ when a watermark should be present in the image, which we denote with $Y_i=1$ for the $i$-th verified image. 
An incorrect prediction ($Y_i=0$) means that either (i) the $p$-value was smaller than $\kappa$ in a non-watermarked image or (ii) it exceeded $\kappa$ for a watermarked image after evasion (line 11).
Finally, the game returns the images after evasion and the evasion rate $e$ (lines 12-13).

The adversary's success is its expected evasion rate $e$ and the visual quality of its images $X^0$, which is commonly measured by the Fréchet Inception Distance (FID)~\cite{heusel2017gans}. 
FID is a perceptual distance computed between two sets of images, and a low FID score indicates a high visual similarity between both sets. 
Similar to the evaluation by Karras et al.~\cite{karras2019style}, we measure the FID between the adversary's images $X$ and the defender's training dataset $D$. 
The adversary's success is a trade-off between the expected evasion rate $e$ and the quality of the images $X^0$.
\begin{align}
    \textrm{Succ}_{\textsc{Evasion}} = \mathbb{E} \left [(e - \textsc{FID}(X^0, D))  \right ]
\end{align}
%
\subsection{Detectability}
%
A detectable watermark poses a threat because it could facilitate an adversary in locating and removing the watermark or spoofing it.
\Cref{alg:watermark-detectability-game} presents the watermark detectability game for a generator $\theta_0$ and the attacker's access to $R_1$ non-watermarked and $R_2$ watermarked images from the generator. 
Our game challenges the adversary to determine the presence of a watermark.

The defender generates a watermarking key (line 2) and embeds a randomly sampled watermarking message into their pre-trained generator $\theta_0$ (lines 3-4).  
The attacker then gets access to $R_1$ images without a watermark and $R_2$ images with a watermark that are samples given random latent codes not controlled by the attacker (lines 7-9). 
A fair coin is flipped, deciding whether the detection attack operates on images containing a watermark, and the attacker has to predict the coin flip (lines 7-8). 
The success of the detectability attack is its expected classification accuracy.  

\begin{algorithm}
\caption{Watermark Detection Game}
\small
\begin{algorithmic}[1]
\game{Detectability}{$\theta_0, R_1, R_2$} 
    \State $\tau \gets \textsc{KeyGen}(\theta_0)$
    \State $m \sim \mathcal{M}$  
    \State $\theta_1 \gets \textsc{Embed}(\theta_0, \tau, m)$
    \State $D_{1} \gets \cup_{i=1..R_1} \{\textsc{Gen}(z,\theta_0)|z\sim \mathcal{Z}\}$  \Comment{No watermark}
    \State $D_{2} \gets \cup_{i=1..R_2}\{\textsc{Gen}(z,\theta_1)|z\sim \mathcal{Z}\}$ \Comment{Watermarked images}
    \State $b \sim \{0,1\}$ \Comment{Unbiased coin flip}
    \State $\hat{b} \gets \mathcal{A}_{\text{Detect}}(\{G(z,\theta_{b})| z\sim \mathcal{Z}\}, D_1, D_2)$ \Comment{Detection attack}
    \State $a \gets [\![\hat{b} = b  ]\!]$ \Comment{Determine correctness}
    \Return $a$
\EndGame
\end{algorithmic}
\label{alg:watermark-detectability-game}
\end{algorithm}
\begin{align}
    \textrm{Succ}_{\textsc{Detection}} = \mathbb{E} \left [ a  \right ]
\end{align}

\textbf{Contrasting with Related Work.} Related work has proposed other methods to measure detectability that sample synthetic images from generators trained on the same dataset with different seeds~\cite{yu2019attributing}. 
However, in these approaches, it is unclear whether the detection was successful because the watermark has been detected or because some other patterns make each generator instance identifiable (e.g., a fingerprint). 
Our notion of detectability can be attributed solely to the impact of the watermark in the synthesized image. 

\section{Conceptual Approach}
This section describes our proposed embedding method for watermarking image generators. 
We describe improvements of our embedding method over existing methods. 
Then, we modify two existing watermarks for GANs to embed into pre-trained generators and propose our improved GAN watermark. 
Finally, we propose three attacks against the robustness of watermarking.  
%
\subsection{Pivotal Tuning Watermarking}
%
Pivotal Tuning Watermarking (PTW) is a method for watermarking a pre-trained generator to preserve latent similarity with the Pivot. 
On input of the same latent code $z\in \mathcal{Z}$, both generators should produce a similar image with imperceptible modifications.
Given a pre-trained generator $\theta_0$ and watermark decoder $\theta_D$, a watermarking message $m\in \mathcal{M}$, a number of iterations $N$, a regularization parameter $\lambda_R$ and a learning rate $\alpha$, PTW returns a watermarked generator $\hat{\theta}$ with high output fidelity to the generator before watermarking for the same latent codes.

\begin{algorithm}
\caption{Pivotal Tuning Watermarking}
\small
\begin{algorithmic}[1]
\Procedure{Embed}{$\theta_0, \tau, m, N, \lambda_R, \alpha$}  
    \State $\tau \gets \textsc{KeyGen}(\theta_0)$ \Comment{Requires a key with which the verification is efficiently optimizable}
    \State $\hat{\theta} \gets \text{copy parameters from } \theta_0$  \Comment{Freeze the Pivot $\theta_0$}
    \For {$i \in \{1..N\}$}  \Comment{\small Embedding loop}
        \State $z \sim \mathcal{Z}$
        \State $x_0 \gets \textsc{Gen}(z, \theta_0)$ \Comment{Frozen Pivot}
        \State $x \gets \textsc{Gen}(z, \hat{\theta})$ \Comment{Watermarked image}
        \State $g_{\hat{\theta}} \gets \nabla_{\hat{\theta}} \mathcal{L}_\text{LPIPS}(x_0,x) + \lambda_R \textsc{Verify}(x, \tau, m)$ 
        \State $\hat{\theta} \gets \hat{\theta} - \alpha \cdot \text{Adam}(\hat{\theta}, g_{\hat{\theta}})$ \Comment{Update the generator}
    \EndFor
    \Return $\hat{\theta}$ \Comment{Watermarked generator}
    \EndProcedure
\end{algorithmic}
\label{alg:ptw-embed}
\end{algorithm}
The watermark decoder neural network extracts messages from images. 
Let $\theta_D$ be a decoder neural network that extracts messages from images, and $m$ is a message that should be embedded. 
Let $\lambda_R$ be the strength of the watermark regularization term, $\alpha$ be the learning rate, and $N$ be the number of steps to optimize the generator $\theta$.  

\Cref{alg:ptw-embed} creates a copy of the pre-trained generator's parameters, called the \emph{Pivot}, and enters a loop (lines 2-3). 
A random latent code is drawn and passed through both variants of the generators to produce two images $x_0, x$ (lines 4-6). 
The loss is computed according to \Cref{eq:pivotal_tuning} where we compute a binary cross-entropy $H$ on the extracted and target messages (line 7). 
The optimization tries to encode as many bits of the message as possible into $x$ while minimizing the LPIPS loss to $x_0$ generated by the pivot.
The model parameters are updated iteratively (line 8), and the tunable generator's parameters and the trained decoder are returned.
The decoder represents the secret watermarking key (line 9). 

\textbf{Overview.} Any image generator that maps from a latent space to images (see \Cref{sec:background}) is compatible with PTW. 
PTW can also be used to embed any image watermarking method~\cite{al2007combined,zhong2020automated} for that can be learned by a watermark decoder network.
We highlight three advantages of PTW over existing embedding methods for GANs~\cite{yu2019attributing, yu2021artificial}. 
\begin{enumerate}\itemsep0mm
    \item \textbf{Speed}: PTW enables watermarking a pre-trained generator up to three orders of magnitude faster than watermarking from scratch. 
    \item \textbf{No Training Data}: PTW requires access only to the generator but not to any training data or the discriminator. 
    \item \textbf{Post-Hoc}: PTW allows embedding watermarks as a post-processing step into any pre-trained generator. 
\end{enumerate}
As stated in \Cref{alg:ptw-embed}, PTW requires access to a differentiable process to extract a message from an image. 
We use a binary message space $\mathcal{M}$ and a deep image classifier $\theta_D$ to extract multi-bit messages from images. 
We propose a \textsc{KeyGen} procedure that generates watermarking keys by leveraging optimization criteria.
Then, we show how existing watermarking methods can be modified to embed them into pre-trained generators efficiently as a baseline for our watermarks.  

\subsection{Generating a Watermarking Key through Optimization}
\label{sec:ptw-keygen}
We now present the implementation of $\textsc{KeyGen}$ that leverages an optimization procedure to derive a watermarking key $\tau$. 
The watermarking key consists of parameters of a deep image classifier $\theta_D$ that can extract multi-bit messages from images and allow backpropagating through the extraction process because the classifier is efficiently differentiable
The ability to efficiently optimize the watermark decoder is central to our approach. 

We generate a key with knowledge of the generator's parameters, which is useful for learning which pixels can be modified easily to hide messages with minimal impact on the visual image quality of the generator.
For example, a generator trained to synthesize faces likely allows encoding more bits per pixel for central pixels in the image that belong to a face, as opposed to pixels at the edge that encode the background. 
The challenge is that none of the existing pre-trained GAN architectures allow the input of a message; hence, the problem becomes how to modulate a message to a generator to subsequently encode this message into its generated images. 

\textbf{Overview.} We use three methods to modulate a message to a generator without modifying its architecture or re-training the generator from scratch. 
All three options are based on training \emph{mapper} neural networks, which map a message to a perturbation of the generator's parameters~\cite{yu2020responsible} or (ii) its inputs~\cite{patashnik2021styleclip} (i.e., the latent codes). 
We implement the parameter mapper by adding the perturbation to a subset of the GAN's weights. 
In our experiments, we modulate $\leq 1\%$ of its total parameters in practice. 
Our parameter-mapper is similar to parameter-efficient fine-tuning (PEFT)~\cite{hu2021lora} techniques and universally applies to any generator architecture.  
Our implementation focuses on modulating convolutional layers of GANs~\cite{zhang2022styleswin}. 

\begin{algorithm}
\caption{Generating Keys through Optimization}
\small
\begin{algorithmic}[1]
\Procedure{KeyGen}{$\theta_0, n, N, \lambda_R$}  
    \State $\theta_P, \theta_Z, \theta_D \gets$ random initialization \Comment{Parameter and latent mappers and message decoder}
    \State $\theta \gets \{\theta_P, \theta_Z, \theta_D\}$  \Comment{$\theta$ denotes all trainable parameters}
    \For {$i \in \{1..N\}$}  
        \State $m \sim \mathcal{M}$ 
        \State $\theta_P \gets \textsc{Mapper}(\theta_P, \theta_0, m, z)$ \Comment{Parameter perturbation}
        \State $\hat{z} \gets \textsc{Mapper}(\theta_Z, \theta_0, m, z)$ \Comment{Latent perturbation}
        \State $x \gets \textsc{Gen}(\theta_0 + \theta_P, \hat{z} + z)$ \Comment{Image with a watermark}
        \State $x_0 \gets \textsc{Gen}(z,\theta_G)$ \Comment{No watermark }
        \State $g_{\theta} \gets \nabla_{\theta} \mathcal{L}_{\text{LPIPS}}(x_0,x) + \lambda_R \textsc{Verify}(x, \theta_D, m)$
        \State $\theta \gets \theta - \alpha \cdot \text{Adam}(\theta, g_{\theta})$ \Comment{update all parameters}
    \EndFor
    \Return $\theta_D$
\EndProcedure
\end{algorithmic}
\label{alg:train-decoder}
\end{algorithm}

We now describe our key generation algorithm that trains a deep image classifier to extract messages. 
Let $\textsc{Mapper}(\theta_0, \theta_1, m, z)$ be a function that takes the parameters of a mapper $\theta_0$, a GAN generator $\theta_1$, a message $m$ and a latent input $z\in \mathcal{Z}$ to predict a perturbation. 
Parameter mappers predict parameter perturbations, and latent mappers predict a perturbation to latents.

\begin{figure}
    \centering
    \includegraphics[width=1\linewidth]{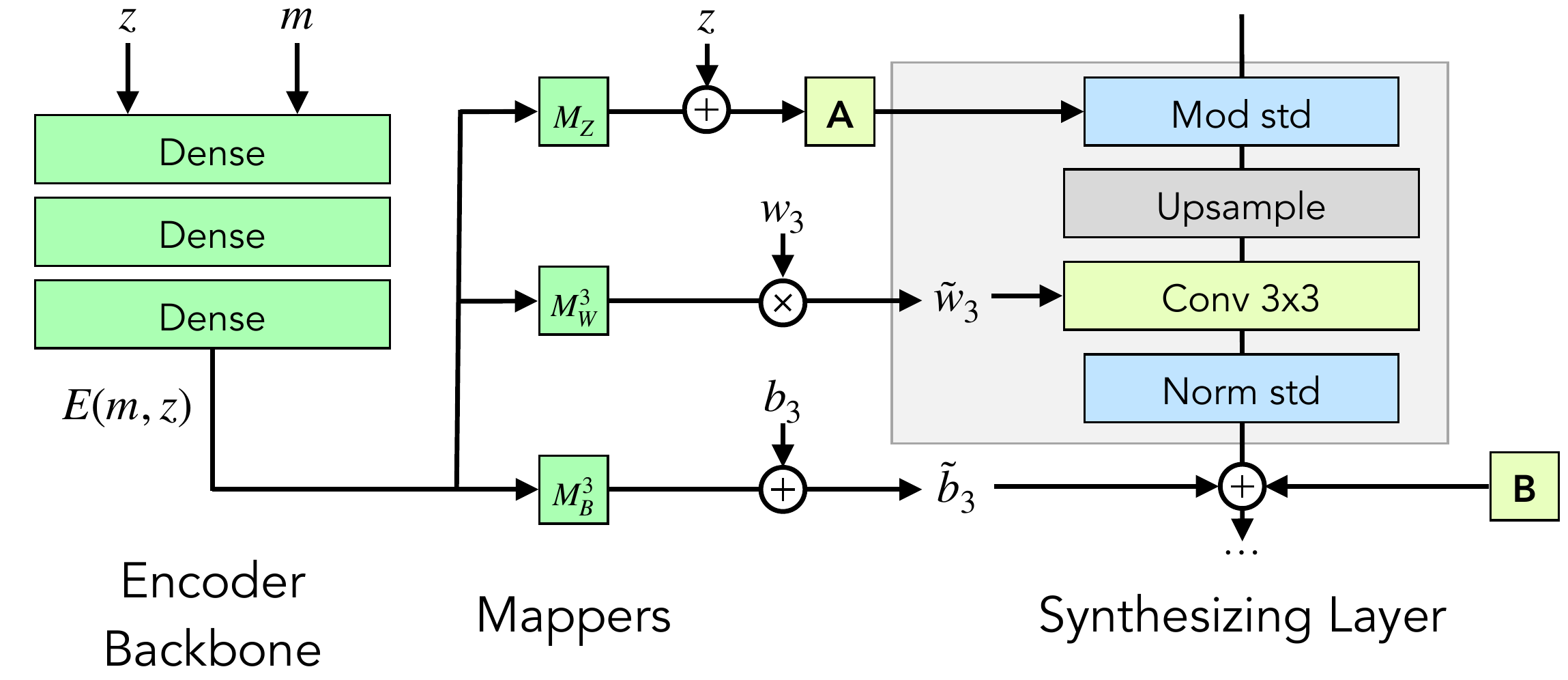}
    \caption{An exemplary illustration of the mappers for a single generator synthesis layer (adapted from \cite{karras2020analyzing}). 
    On input of a latent code $z$ and a message $m$, the mappers (in green) modulate the generator's weights and inputs. \framebox{A} is an affine transform and \framebox{B} is random noise sampled during inference. }
    \label{fig:ptw-mapper}
\end{figure}

\textbf{Training the Decoder.} \Cref{alg:train-decoder} presents pseudocode for the key generation procedure. 
First, we randomly initialize a parameter mapper $\theta_P$, a latent mapper $\theta_Z$, and a message decoder $\theta_D$ (lines 2-3). 
Then, we iteratively optimize all trainable parameters by randomly sampling a message and predicting a parameter and a latent perturbation (lines 5-7).
We generate a watermarked and non-watermarked image and calculate the gradient on the loss between the quality difference and the verification loss (lines 8-10). 
Finally, we update all trainable parameters (line 11), and at the end of the optimization, we return the parameters of the decoder (line 12). 
\Cref{fig:ptw-mapper} illustrates the modulation of a message to a synthesizing layer in a generator. 

The returned decoder can extract messages from any image. 
Notably, the decoder learns an encoding of the message that causes the least degradation in visual image quality (according to the LPIPS loss). 
We refer to \Cref{fig:teaser}, where the most perturbed pixels are pixels with high semantic value, such as the eyes and nose of a face. 
Since the decoder depends on the generator instance, a different decoder should be trained for each instance for the best visual quality, but decoders can be re-used if the model architectures are similar. 
Next, we describe modifications to existing watermarks~\cite{yu2019attributing,yu2020responsible} that allow embedding them into pre-trained generators. 

\subsection{Modifying Existing Watermarks}
\label{sec:modifying-existing-watermarks}
Two existing watermarks for GANs require re-training the generator from scratch to embed a watermark~\cite{yu2019attributing,yu2021artificial}.  
This section describes modifications to both methods for watermarking pre-trained generators as a baseline comparison to our method.
First, we briefly summarize both watermarking methods and then describe our modifications to enhance their efficiency. 

\textbf{Summary.} The first watermark, which we call \texttt{Yu1}~\cite{yu2019attributing}, trains an encoder-decoder network on real images by marking the images with imperceptible patterns. 
The GAN is trained from scratch on the marked training data. 
The second watermark, which we call \texttt{Yu2}~\cite{yu2021artificial}, modifies the GAN's training objective and is trained from scratch. 
Some of their modifications were invented to mitigate training instabilities such as mode collapse~\cite{thanh2020catastrophic} or consistency losses, which are problems that do not appear during fine-tuning. 
The authors modulate the watermark with a parameter mapper $\theta_P$ and embed the watermark by adding the predicted parameter perturbation to the generator's weights. 
We find that embedding without PTW has a substantial impact on the generator's image quality and that using PTW can preserve a much higher image quality.
We refer to the authors' papers for a more detailed description of their works. 

\textbf{Modifications.} The required modifications for \texttt{Yu1} are straightforward: Instead of training on real training data, we stamp synthetic data and use their decoder network for embedding a watermark with PTW.
For \texttt{Yu2}, we ignore all additional losses that address training instabilities or consistency losses and train their weight mapping network instead via fine-tuning on synthetic data while freezing the generator's weights. 
We embed the \texttt{Yu2} watermark using the author's approach of summing the weight mapper's prediction to the generator's parameters. 
\subsection{Attacks to Evade Watermark Detection}
\label{sec:attacks-against-robustness}
\begin{figure}
    \centering
    \includegraphics[width=1\linewidth]{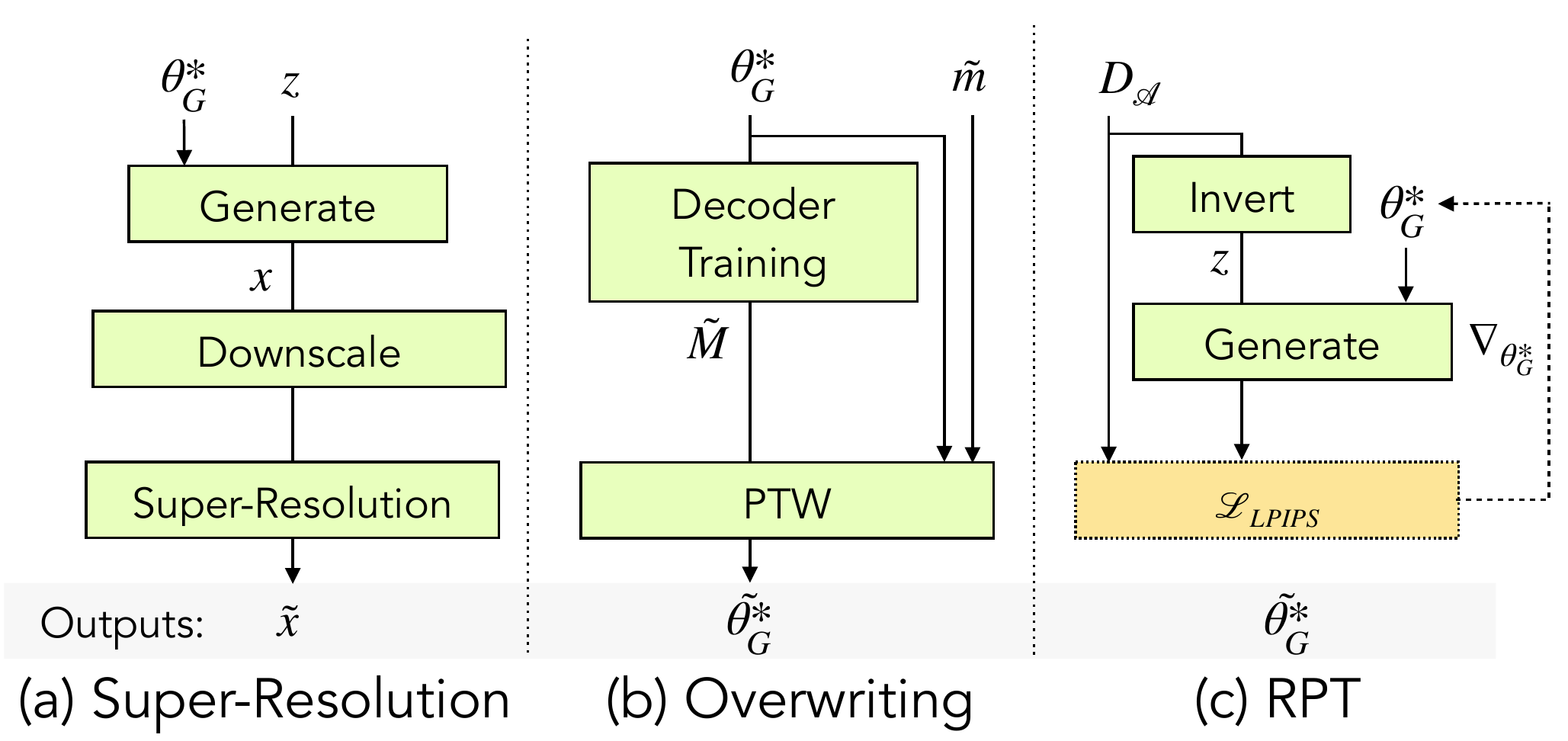}
    \caption{An illustration of our three attacks against the robustness of generator watermarking. RPT stands for Reverse Pivotal Tuning. The function \textsc{Invert} maps images to latent codes, and \textsc{Downscale} reduces the resolution of an image.  }
    \label{fig:attacks}
\end{figure}
This Section proposes three novel adaptive attacks against the robustness of model watermarking for image generators. 
Recall from \Cref{sec:threat_model} that we consider two adversaries: a black-box and a white-box adversary.
Previous work~\cite{yu2019attributing, yu2021artificial} assumes only a black-box attacker who can execute any of these five attacks: blurring, cropping, image noising, JPEG compression, or quantization. 
We refer to \Cref{appendix:sec-attack-description} for a detailed description of all attacks and parameters, including those from previous work.
\Cref{fig:attacks} illustrates our proposed black box and two white-box attacks. 

\subsubsection{Black-box Attacks } 
Our black-box attacker first scales the resolution of an image down by a factor $\rho$ and then uses \emph{super-resolution} models~\cite{rombach2022high} to upscale the image to its previous resolution. 
A super-resolution model can upscale images by interpolating details that are not sharp in the low-resolution image. 
Such super-resolution models enable an attacker to apply stronger perturbations to images in an attempt to remove their watermark with a smaller impact on the image quality.
Super-resolution models have been demonstrated to generalize well to out-of-domain data, meaning the attacker does not need access to the generator's training data. 
While our attacks use pre-trained models from related work~\mbox{\cite{rombach2022high}} to achieve super-resolution, we are the first to apply super-resolution models to undermine watermarking in image generators.
Previous attacks~\cite{yu2019attributing,yu2020responsible,yu2021artificial} use image augmentation techniques such as blurring or noising to remove the watermark.  

\subsubsection{White-box Attacks} 
We propose two adaptive white-box attacks called \emph{Overwriting} and \emph{Reverse Pivotal Tuning} (RPT). 
In the overwriting attack, the attacker trains their own watermark decoder (see \Cref{alg:train-decoder}) and then uses PTW for watermarking the generator using a random message. 
The success of the overwriting attack depends on the similarity between the defender's watermarking key $\tau$ and the attacker's watermarking key $\tilde{\tau}$. 
The overwriting attack can successfully remove the watermark if both keys modulate similar pixels in the input.

\begin{algorithm}
\caption{(White-box) Reverse Pivotal Tuning (RPT)}
\small
\begin{algorithmic}[1]
\Procedure{Invert}{$x, \hat{\theta}$} 
    \Return $\underset{z\in \mathcal{Z}}{\argmin}~\mathcal{L}_{\text{LPIPS}}(\textsc{Gen}(\hat{\theta}, z), x)$
\EndProcedure
\end{algorithmic}
\begin{algorithmic}[1]
\Procedure{RPT}{$\hat{\theta}, D, N, \alpha$}
    \State $Z \gets \{\textsc{invert}(x, \hat{\theta})|x\in D\}$ 
    \For {$i \in \{1..N\}$}  
        \State $g_{\hat{\theta}} \gets \nabla_{\hat{\theta}} \mathcal{L}_{\text{LPIPS}}(\textsc{Gen}(\hat{\theta}, Z_i), D_i)$
        \State $\hat{\theta} \gets \hat{\theta} - \alpha \cdot \text{Adam}(\theta, g_{\hat{\theta}})$
    \EndFor
    \Return $\hat{\theta}$
\EndProcedure
\end{algorithmic}
\label{alg:rpt}
\end{algorithm}
\Cref{alg:rpt} implements our RPT attack.
The attacker has access to a watermarked generator $\theta^*_G$, a limited set of $R$ non-watermarked, real images $D^R$ and performs the RPT attack for $N$ steps with a learning rate $\alpha$. Their goal is to regularize the generator to synthesize images that are visually similar to their real images (i.e., they have high visual quality), but do not retain a watermark. 
The RPT attack consists of two stages: (1) inversion of the real images (line 5) and (2) Pivotal Tuning so that the inverted images have a high visual similarity to the real images (lines 5-6). 
RPT should be successful with the availability of many non-watermarked images. 

\section{Experiments}
We describe our experimental setup and specify measured quantities, namely capacity, utility, detectability, and robustness. 
Then, we measure detectability and robustness (see \Cref{sec:threat_model}) and compare our watermarking method to the modified methods from related work (see \Cref{sec:modifying-existing-watermarks}). 

\begin{figure*}
    \centering
    \includegraphics[width=1.\linewidth]{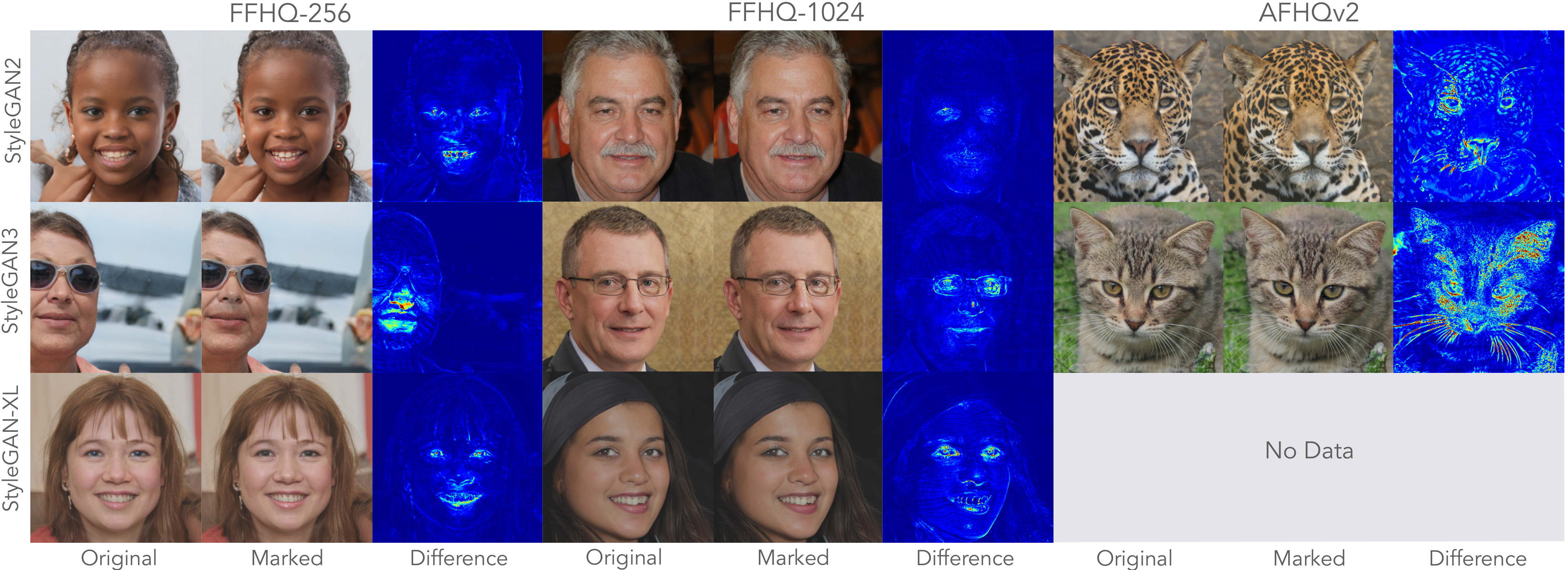}
    \caption{Images synthesized using our watermarked generators on different datasets and model architectures. We show the image synthesized by the generator (i) before and (ii) after watermarking, and (iii) the difference between the watermarked and non-watermarked images. StyleGAN-XL does not provide a pre-trained model checkpoint for AFHQv2.}
    \label{fig:watermark-demo}
\end{figure*}

\subsection{Experimental Setup}
\textbf{Datasets.} We experiment with three datasets for which pre-trained, high-quality generators have been made publicly available. 
FFHQ~\cite{karras2019style} consists of 70k human faces in various poses. 
We experiment with a lower-resolution version of the dataset at $256^2$ pixels, which we refer to as FFHQ-256, and the high-quality version FFHQ at $1\,024^2$ pixels.
In addition, we experiment with AFHQv2~\cite{choi2020stargan}, which consists of roughly 16k animal images and has a resolution of $512^2$ pixels.

\textbf{Pre-Trained Generators.} We experiment with three StyleGAN-based architectures: StyleGAN2~\cite{karras2019style}, StyleGAN3~\cite{karras2021alias} and StyleGAN-XL~\cite{sauer2022stylegan}. 
We select the StyleGAN architectures because (1) this architecture achieves state-of-the-art FID values on the surveyed datasets and (2) many high-quality model checkpoints are publicly available that have been trained with different seeds\footnote{\url{https://github.com/NVlabs/stylegan3}}\footnote{\url{https://github.com/autonomousvision/stylegan-xl}}. 
State-of-the-art image generator architectures are evaluated using the same checkpoints that we are using.
Therefore, the image quality of our watermarked generated images can be compared to the image quality in ongoing research on image generation models. 

\textbf{Framework.} We implement all watermarking methods from scratch in PyTorch 1.13. 
While implementations for \texttt{Yu1}\footnote{\url{https://github.com/ningyu1991/ArtificialGANFingerprints}} and \texttt{Yu2}\footnote{\url{https://github.com/ningyu1991/ScalableGANFingerprints}} exist, we could not reproduce their results with the provided implementation. 
\texttt{Yu1} never converges, and \texttt{Yu2} is implemented in Tensorflow version 1, which is no longer supported by modern GPUs, meaning we cannot reuse their source code or load the provided generator checkpoints. 

\subsection{Metrics}
\textbf{Utility.} Similar to existing work~\cite{karras2019style}, we measure the utility of a generator by its Fréchet Inception Distance (FID)~\cite{heusel2017gans}. 
Lower FID indicates a higher utility. 
Like Karras et al.~\cite{karras2019style}, we measure FID between $50,000$ generated and real images. 
For AFHQv2, we use only $16,000$ real images due to the limited dataset size. 

\textbf{Capacity.} We measure the capacity of a watermark in bits by the difference in the expected number of correctly extracted bits from watermarked and non-watermarked images. 
The expected rate of correctly extracted bits equals $0.5$ for non-watermarked images, assuming messages are randomly sampled. 
Let $m\in \{0,1\}^n$ be a message, $\tau$ the secret watermarking key, and $\theta$ are the parameters of a generator.
The capacity of the generator is computed as follows.
\begin{align}
    C_\theta =  n \cdot \underset{z \in \mathcal{Z}}{\mathbb{E} }\Big[ \textsc{BER}(\textsc{Extract}(G(z;\theta), \tau), m)- 0.5 \Big] 
\end{align}
The function \textsc{BER} computes the bit error rate between messages.
It is straightforward to achieve a high capacity by overwriting a significant portion of the image. However, doing so also decreases the image's quality, which can be measured and visualized as the capacity/utility trade-off. 

\textbf{Decision Threshold.} We consider a watermark to be \emph{removed}, if we can reject the null hypothesis $H_0$ with a $p$-value less than $0.05$. 
The null hypothesis states that $k$ matching bits were extracted from the synthetic images by random chance. 
Quantitatively, the probability of this event is calculated as $\text{Pr}(X>k|H_0) = \sum_{i=k}^n \binom{n}{i} 0.5^n$.
In practice, for a watermark with $n=40$ bits, we need to extract at least $26$ bits correctly, meaning that we verify the presence of a watermark by correctly extracting $C_\theta \geq 6$ in bits. 

\subsection{Runtime Analysis}

\begin{table}
    \small
    \begin{center}
        \begin{adjustbox}{max width=\linewidth}
        \begin{tabular}{lccc}
        \hline
        Model & StyleGAN2 & StyleGAN3 & StyleGAN-XL \\
        \hline
        FFHQ-256 & 158h & 482h & 552h \\\rowcolor{gray!13}
        FFHQ-512 & 384h & 662h & 1285h \\
        FFHQ-1024 & 929h & 1161h & 1456h \\
        \hline
        \end{tabular}
        \end{adjustbox}
    \end{center}
\caption{Time measured in GPU hours required for training generators \emph{without} watermarking (from scratch) on FFHQ~\cite{karras2019style} on 8xA100 GPUs.}
\label{tab:speed-up-ptw}
\end{table}

To calculate the speed-up of PTW over other methods~\mbox{\cite{yu2020responsible,yu2021artificial}}, we compare it with training non-watermarked generators from scratch. 
This comparison is fair, as watermarking is not expected to decrease a generator's training time.
We estimate the total runtimes in GPU hours using the suggested parameters in the relevant GAN  papers~\mbox{\cite{karras2020analyzing, karras2021alias, sauer2022stylegan}} on 8xA100 GPUs.

\mbox{\Cref{tab:speed-up-ptw}} shows the estimated training runtimes from scratch for each generator on FFHQ~\mbox{\cite{karras2020analyzing}} at varying pixel resolutions. 
For instance, training a StyleGAN-XL model on FFHQ at a resolution of $256^2$ pixels requires 552 GPU hours. 
With PTW, watermarking a pre-trained generator on FFHQ requires only about 0.5 GPU hours, which is three orders of magnitude improvement for high-resolution generators. 
Our approach also requires training the watermarking decoder (see \mbox{\Cref{alg:train-decoder}}), which is a one-time upfront cost of about 2 GPU hours.

\subsection{Capacity/Utility Trade-off}

\begin{figure*}%
    \centering
    \subfloat[FFHQ-256.\label{fig:ffhq256-capacity-utility}]{{\includegraphics[width=.33\linewidth]{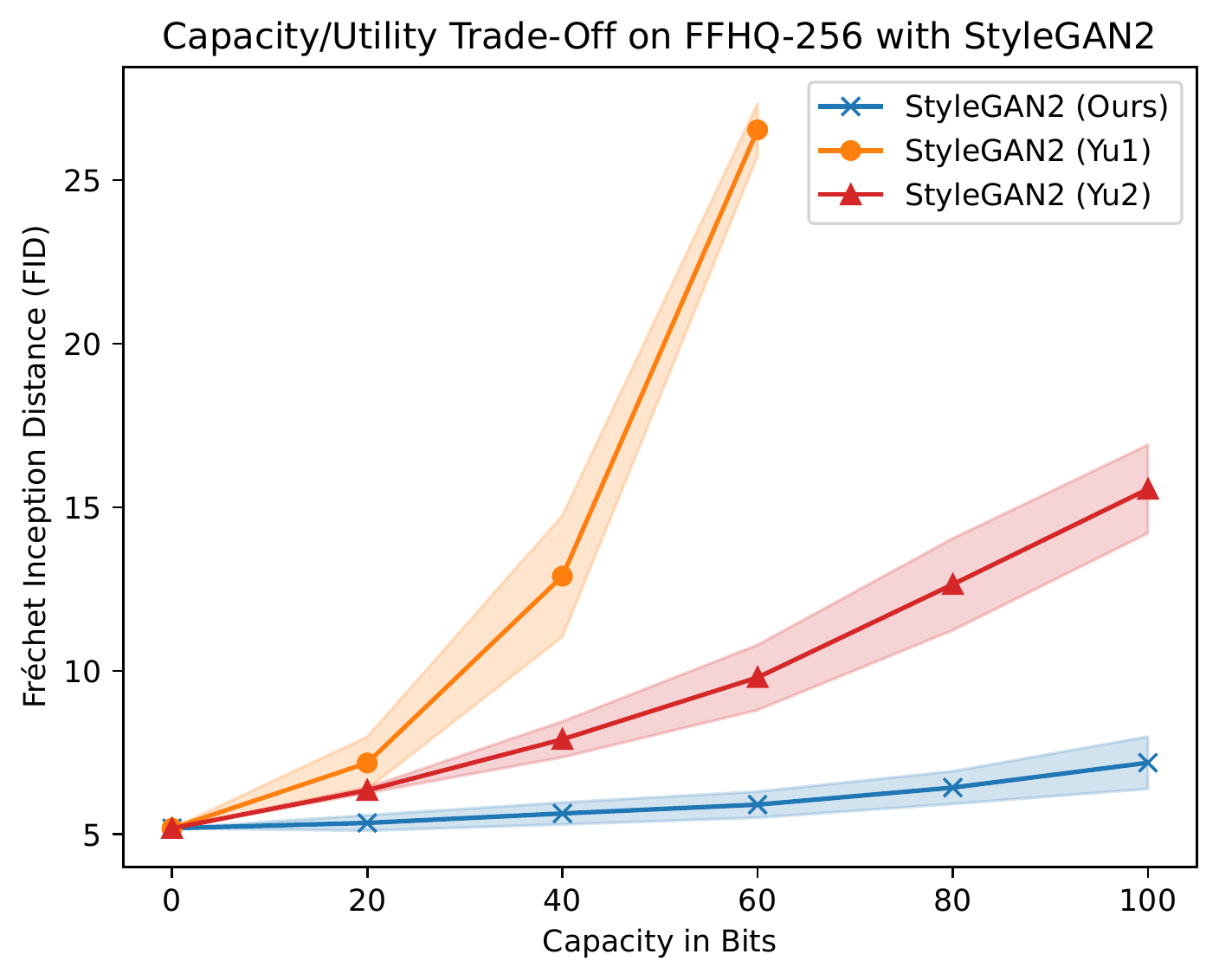} }}
    \subfloat[FFHQ-1024. \label{fig:ffhq1024-capacity-utility}]{{\includegraphics[width=.33\linewidth]{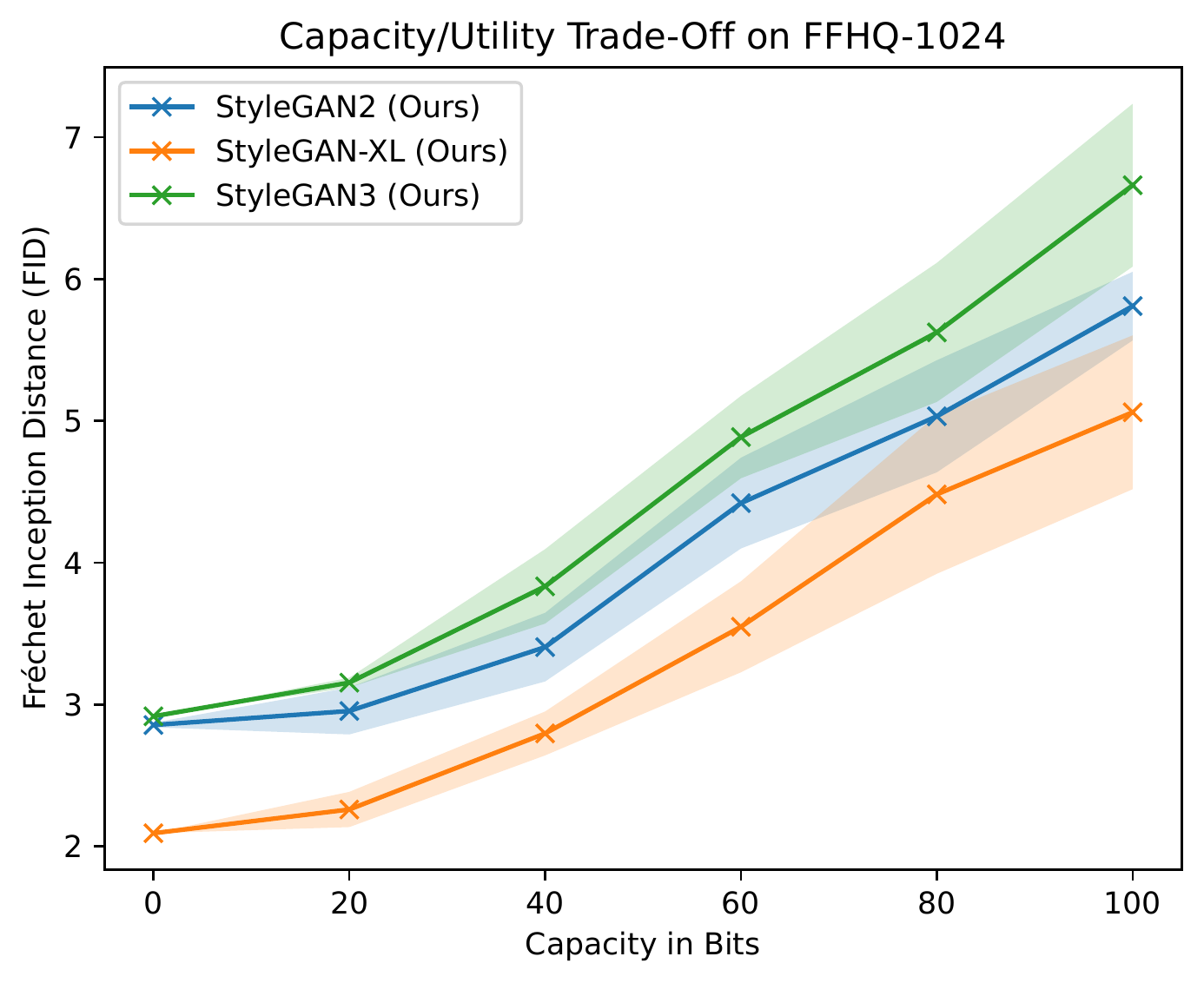} }}
    \subfloat[AFHQv2-512. \label{fig:afhq512-capacity-utility}]{{\includegraphics[width=.33\linewidth]{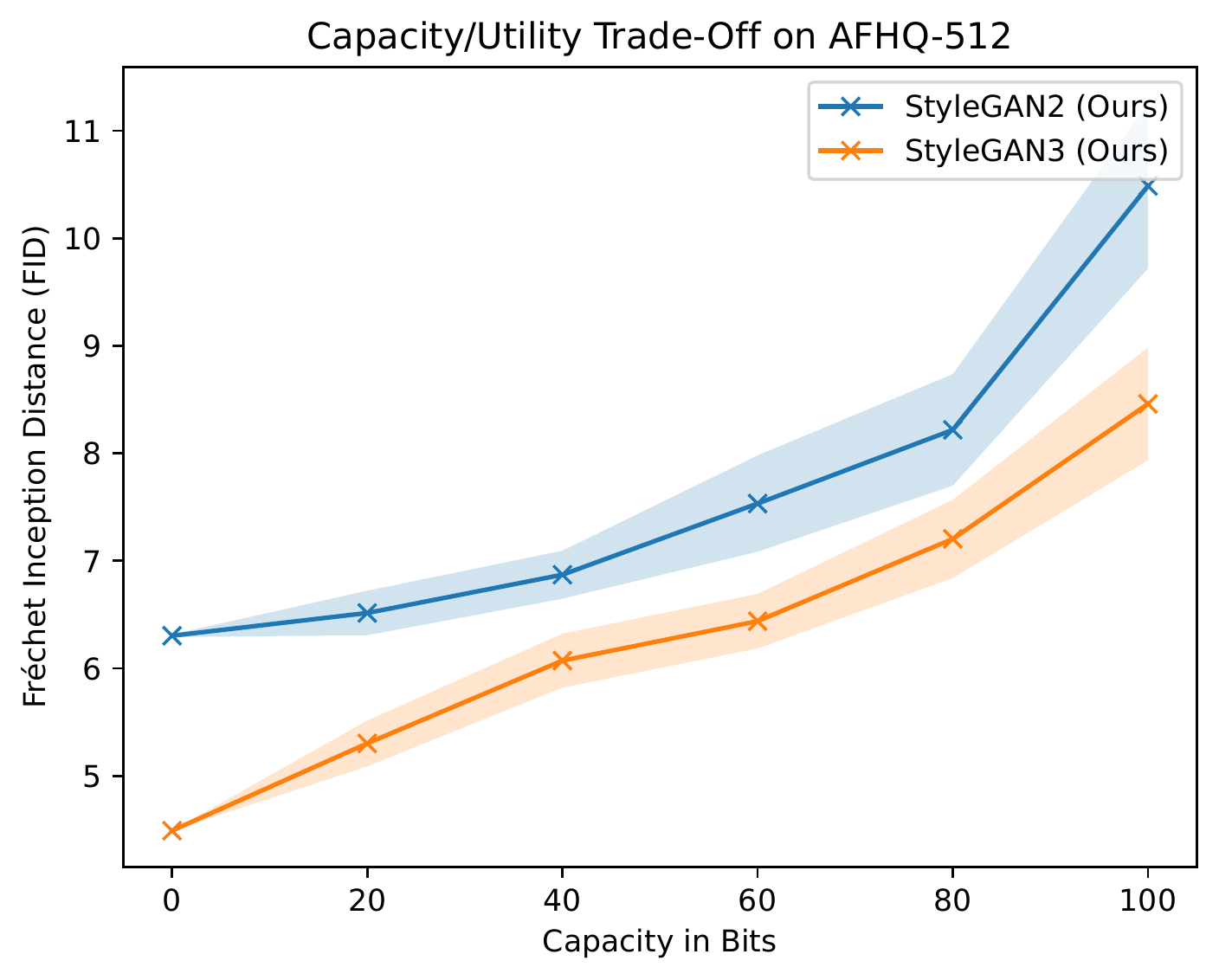} }}
    \caption{The capacity/utility trade-off. \Cref{fig:ffhq256-capacity-utility} shows our watermark compared to two existing, modified watermarks for StyleGAN2.
    \Cref{fig:ffhq1024-capacity-utility} shows our watermarking for on FFHQ-1024 using three different generator architectures. \Cref{fig:afhq512-capacity-utility} shows our watermark on a domain other than faces (wild animals) using different generator architectures. 
    The shaded area illustrates the standard deviation for three repetitions.}%
\end{figure*}
This section summarizes our results on the capacity/utility trade-off on various datasets, model architectures, and compared to two existing, modified watermarks: \texttt{Yu1} and \texttt{Yu2}.

\textbf{Visual Inspection.} \Cref{fig:watermark-demo} shows images synthesized by our watermarked generators on all three surveyed datasets. 
The columns show the original image synthesized before watermarking the image synthesized after watermarking, and their differences in the form of a heatmap.  
Heavily modified regions are highlighted in yellow and red. 
In both versions of the facial image datasets, we observe that our watermark focuses on pixels located on the face of the generated person, most prominently its eyes.
Upon closer inspection, the network modifies the eyes and mouth area of a face strongest and is invariant to the location of the face in the image. 
For AFHQv2, we observe that the pixels are more spread out onto the entire image. 
In the next subsection, we compare our watermark quantitatively to other existing watermarks. 

\textbf{Comparison to Existing Watermarks.} 
\Cref{fig:ffhq256-capacity-utility} shows the trade-off on FFHQ-256 for different watermarking methods using a StyleGAN2 architecture. 
We plot the embedded bits $C_{\theta}$ against the FID.  
Our method outperforms the other two approaches substantially, as we can embed 100 bits with a similar loss in utility as embedding 20 bits using \texttt{Yu2}. 
In contrast, \texttt{Yu1} is not competitive, even though it employs PTW as its embedding strategy. 
Upon analyzing the generated images, we observe that \texttt{Yu1} is not sensitive to the capacity per pixel and attempts to encode many bits of the message into background pixels, which noticeably deteriorates the generator's quality. 
We believe this method works better when the entire generator is re-trained from scratch, as the generator can learn to allocate capacity to arbitrary pixels. 
For FFHQ-256, using our watermark, encoding 40 bits only worsens the FID by approximately 0.3 points.   

\textbf{Watermarking High-Quality Generators.} \Cref{fig:ffhq1024-capacity-utility} shows the trade-off using our watermark across three generator architectures on FFHQ-1024. 
Compared to FFHQ-256, which has a FID of over 5, the generators trained on FFHQ-1024 have a much lower FID of less than 3. 
Our watermark embeds up to 40 bits with little loss in utility, but the FID deteriorates quickly when embedding more than 40 bits.
For StyleGAN2, we measure a FID deterioration of almost 3 points when embedding 100 bits. 
We believe the effect on the FID is greater for the high-quality dataset due to two factors: (1) high-quality images with a low FID may be more sensitive to modifications, and (2) our watermark decoder downscales images to $224^2$ pixels, which means that our decoder cannot extract more information from larger images. 
Our decoder is a ResNet18~\cite{he2016deep} model designed for this resolution.
Nonetheless, we demonstrate that watermarking high-quality generators is possible using our method.

\textbf{Watermarking Different Domains.} \Cref{fig:afhq512-capacity-utility} shows the capacity/utility trade-off across two generator architectures for the domain of animal images.
We cannot evaluate StyleGAN-XL on AFHQv2 because no pre-trained checkpoint was available for this dataset. 
We aim to demonstrate that our watermark is not restricted to just the facial image domain. 
\Cref{fig:afhq512-capacity-utility} shows that our approach can embed watermarks up to 100 bits, although we observe a strong deterioration in FID of more than 4 points, at which point the watermark is (barely) visually perceptible.
While it is possible to embed 100 bits, given our results, we believe that 40 bits are more practically relevant as the deterioration in FID is less than one point for StyleGAN2, and the watermark is not easily perceptible. 
Interestingly, the deterioration in FID is stronger for the animal domain, which we attribute to a larger output diversity.
AFHQv2 contains images of multiple different animal species in diverse poses.

\subsection{Detectability}

\begin{figure*}%
    \centering
    \subfloat[]{{\includegraphics[width=.33\linewidth]{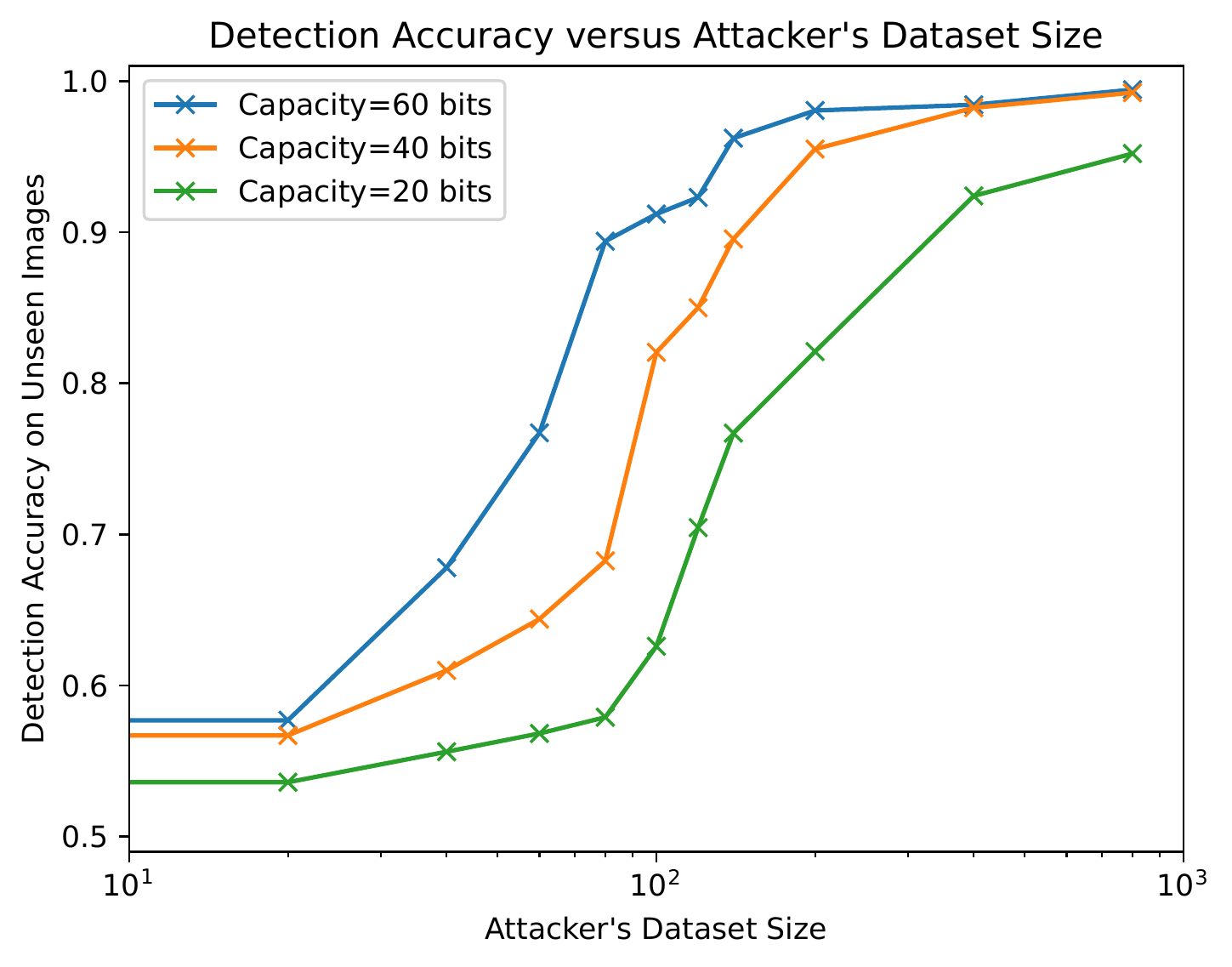} 
    \label{fig:detectability}}}
    \subfloat[]{{\includegraphics[width=.33\linewidth]{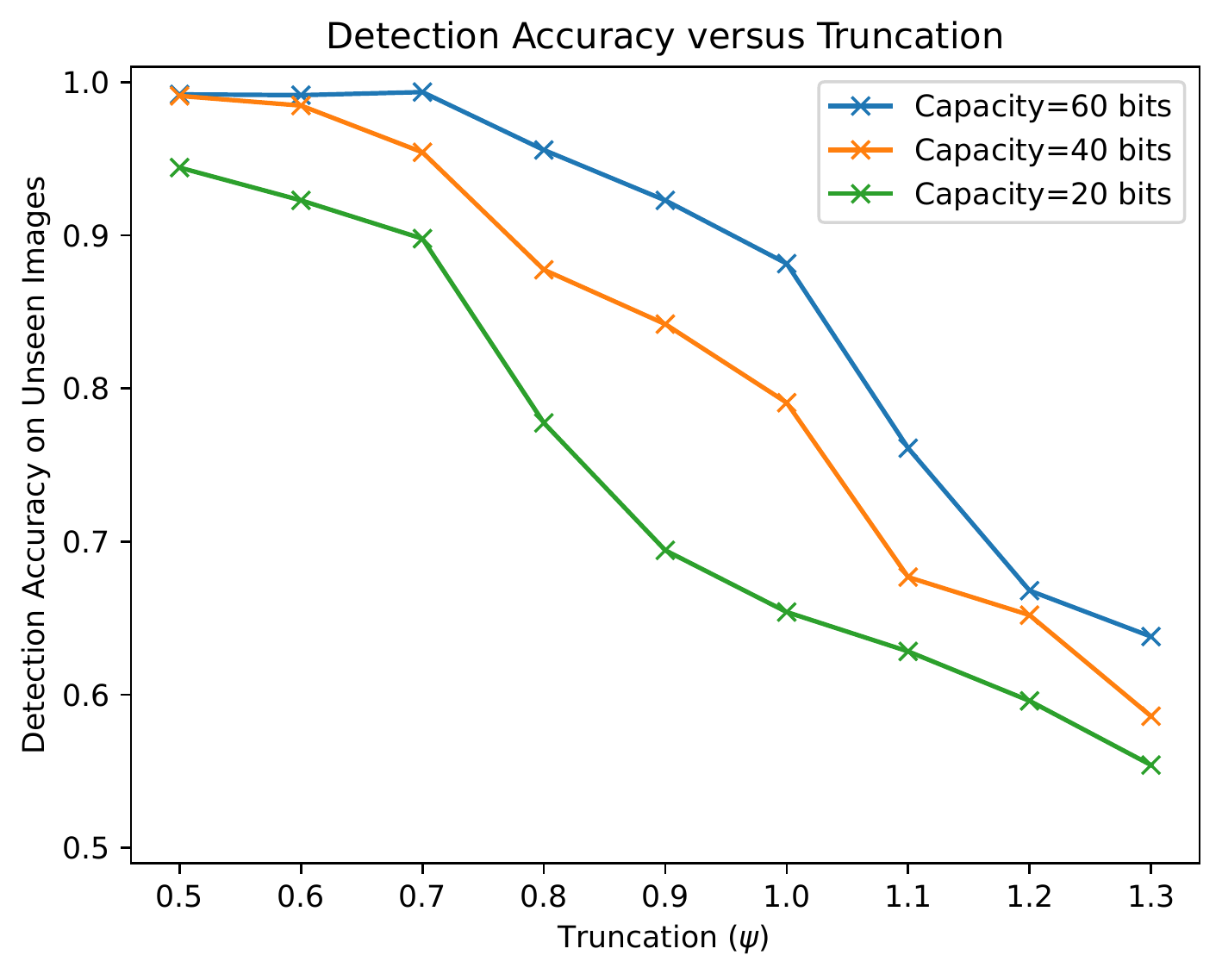} \label{fig:detection-truncation}}} 
    \subfloat[]{{\includegraphics[width=.33\linewidth]{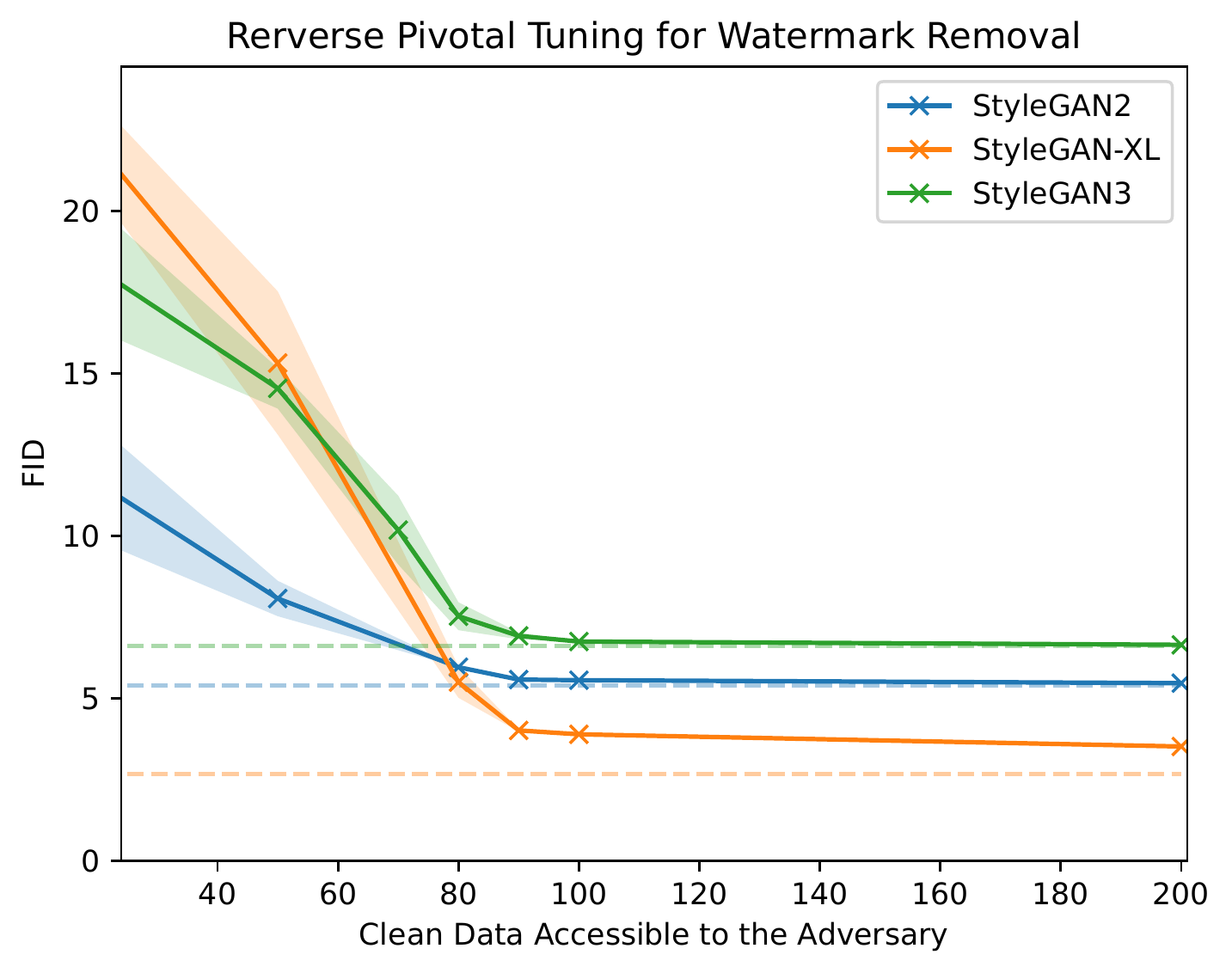} \label{fig:rpt-ablation} }}
    \caption{(a) The detection accuracy in relation to the adversary's dataset size for different capacities without truncation. (b) The detection accuracy plotted against truncation when fixing the adversary's dataset size to $\leq100$ labeled images. (c) An ablation study for the number of real, non-watermarked data used during our adaptive Reverse Pivotal Tuning attack. The shaded areas denote the standard deviation (N=3), and the dashed horizontal lines show the generator's FID before the attack.  }%
\end{figure*}

Our first experiment measures the detectability of our watermark at different capacities. 
We recall from our security game in \Cref{alg:watermark-detectability-game} that the attacker can access $R_1, R_2$ non-watermarked and watermarked images. 
In our experiments, we set $R_1=R_2$ and evaluate the detectability of images synthesized by generators that have been watermarked with varying capacities.
We use a standard, pre-trained ResNet-18 and fine-tune it for the detection task using an Adam optimizer. 

\textbf{Detectability versus Dataset Size. }  
\Cref{fig:detectability} shows the detection accuracy $p$ of our attack against the number of labeled images available to the adversary. 
As expected, a higher watermark capacity results in a higher detectability of watermarked images. 
We observe that an adversary with access to $\geq 400$ labeled images has a classification accuracy of over $90\%$ in classifying watermarked/non-watermarked images for any capacity. 
An adversary with access to $\leq 100$ images cannot reliably detect watermarked images if the capacity is at most $40$ bits.
Our results show that the detection algorithm cannot detect our watermark unless the attacker can access a relatively large set of labeled, non-watermarked images. 
Next, we evaluate the influence of the latent code's sampling strategy on the watermark detectability. 

\textbf{Truncation Trick.} Generating an image from a GAN requires sampling a latent code. 
The \emph{truncation trick}~\cite{brock2018large} is a technique used for generative models to limit the range of values for a latent code and allows for controlling the diversity and quality of the generated images. 
If the truncation threshold $\psi$ is low, samples will have a high similarity to the real training data but limited diversity, meaning they may appear similar to each other.
We identify that truncation plays a significant role in the ability of an adversary to detect watermarks. 
\Cref{fig:detection-truncation} shows that the detection accuracy decreases with an increasing diversity of the generator's output when fixing the number of samples available to the adversary. 

\subsection{Robustness}
We evaluate the watermark's robustness against several types of attacks, including (1) black-box attacks like cropping, blurring, quantization, noising, JPEG compression, and our super-resolution attack, and (2) two white-box attacks, overwriting, and Reverse Pivotal Tuning (see \Cref{sec:attacks-against-robustness}).
We refer to \Cref{appendix:sec-attack-description} for describing the attacks and parameters we use during our evaluation. 
All attacks are evaluated against generators that have been watermarked with a capacity of at least $C_\theta\geq 40$ bits. 
Embedding 40 bits only deteriorates the generator's FID by about $0.3$ points on average for StyleGAN2 on FFHQ. 

\subsubsection{Black-box Attacks}
\textbf{Latent Space Analysis.} We examine whether there are points in the generator's latent space that synthesize high-quality images without a watermark. 
If such points exist, an attacker could attempt to find them and sample the generator on these points. 
We test for such latent subspaces using three sampling methods: (i) truncation, (ii) latent interpolation, and (iii) style-mixing (for the StyleGAN architectures). 
Truncation restricts the distance of a latent code to the global average. 
Latent interpolation samples point on a line between latent codes, and style-mixing combines intermediate latent codes $w\in \mathcal{W}$ and feeds them to the generator~\cite{karras2019style}. 
Our results show that the mean capacity remains unaffected by the sampling method, meaning we were able to successfully extract the watermark message in all cases.
We conclude from these results that the watermark generalizes to the generator's entire latent space. 

\begin{figure}
    \centering
    \includegraphics[width=1\linewidth]{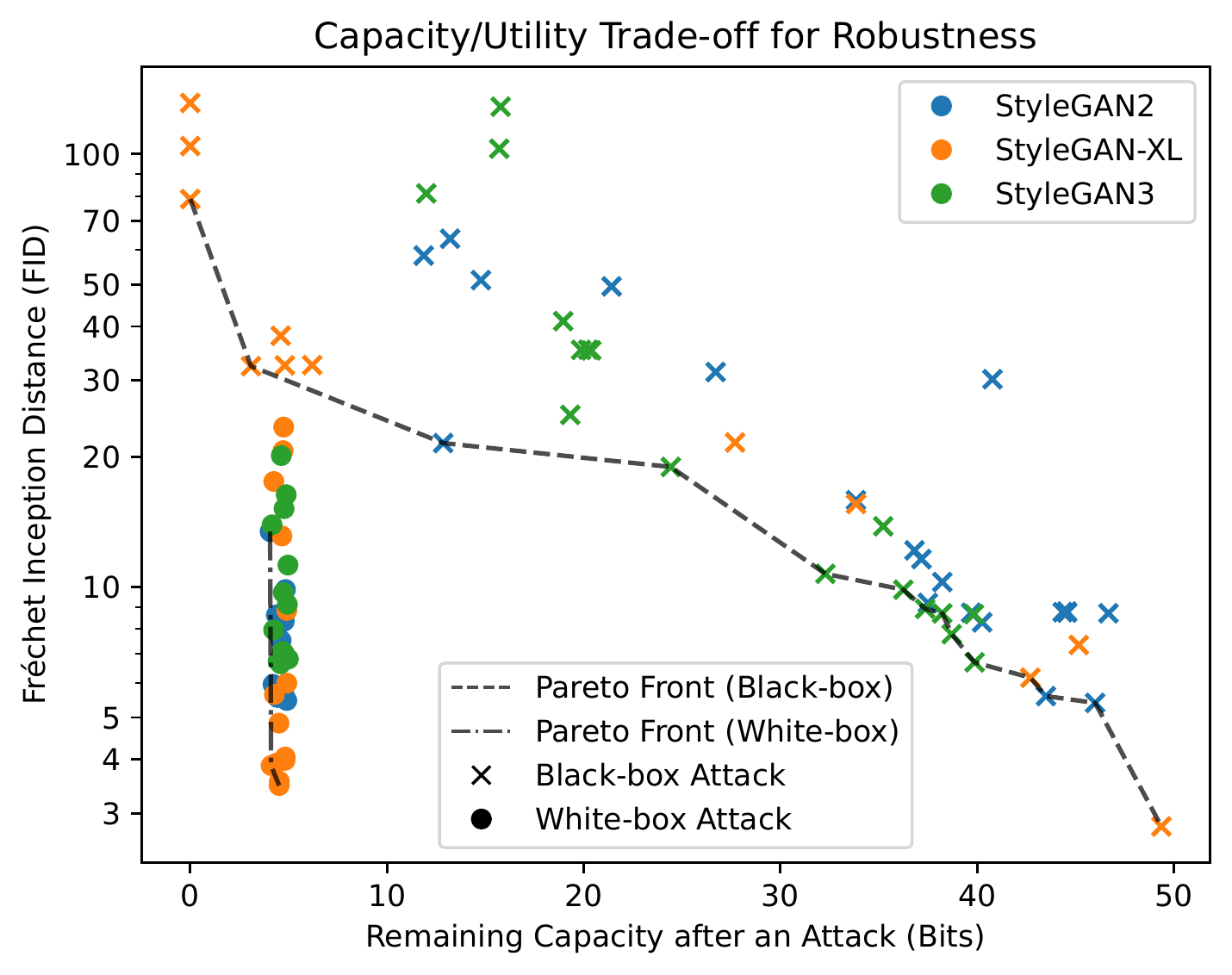}
    \caption{
    This Figure shows the robustness of our watermark against all surveyed attacks. We highlight black-box and white-box attacks that are in the Pareto front. 
    }
    \label{fig:robustness-scatter}
\end{figure}

\textbf{Removal Attacks.} Next, we perform all surveyed removal attacks against all watermarked generators and measure the evasion rate and FID with $K=50,000$ synthetic images. 
A summary of all black-box attacks is shown as a scatter plot in \Cref{fig:robustness-scatter}. 
The Figure shows the remaining capacity after an attack on the x-axis and utility (measured by the FID) on the y-axis. 
The Pareto front, which represents the optimal trade-off between capacity and utility, is highlighted and represents the best attack out of all surveyed attacks that an adversary could choose. 
We find that none of the black-box attacks are effective at removing a watermark, but our super-resolution attack is always part of the Pareto Front. 

The Pareto front represents the best capacity/utility trade-off a black-box attacker can achieve using these attacks. 
For example, a black-box attacker can reduce the capacity by 10 bits from 50 to 40 but, in doing so, reduces the FID by over 6 points. 
Our super-resolution attack is on the Pareto front, but cannot remove the watermark. 
Removal is only possible when the FID drops to 30, at which point the image quality has been compromised.
\Cref{tab:robustness-table} summarizes the best-performing black-box attacks for the three evaluated generator architectures. 
Each attack has a single parameter that we ablate over using grid search.
We refer to \Cref{appendix:sec-attack-description} for a detailed description of all attacks and parameters we used in this ablation. 
\Cref{tab:robustness-table} lists those data points that either remove the watermark ($C_\theta < 5$) or, if the watermark cannot be removed, the data point with the lowest FID. 
None of the black-box attacks, including our super-resolution attack, successfully remove the watermark while preserving the generator's utility.

\subsubsection{White-box Attacks}

\textbf{Overwriting.}
 \Cref{tab:robustness-table} shows that overwriting can remove watermarks but deteriorates the generator's image quality, measured using FID, by approximately 3 points for StyleGAN2 and 6 points for StyleGAN-XL.
Such a deterioration in FID likely prevents attacks in practice because low-quality deepfakes are more easily detectable. 
Overwriting implicitly assumes knowledge of the defender's watermarking method, which may not be true in practice. 
Overwriting could cause a noticeable decline in FID if the attacker's and defender's watermarking methods differ. 

\textbf{Reverse Pivotal Tuning} is substantially more effective than the overwriting attack as it preserves the FID of the generator to a greater extent.
We found that an attacker accessing 200 real, non-watermarked images can remove any watermark without causing a noticeable deterioration in FID. 
This means that with access to less than 0.3\% of the training dataset, a white-box adversary can remove any watermark.
In the case of StyleGAN-XL, using 200 images leads to a decrease in FID of less than one point (from 2.67 to 3.52). 

\begin{table}[]
    \centering
    \small
    \begin{tabular}{@{}lcccccc@{}}
\toprule
                                & \multicolumn{2}{l}{StyleGAN2} & \multicolumn{2}{l}{StyleGAN-XL} & \multicolumn{2}{l}{StyleGAN3} \\ \midrule \rowcolor{gray!13}
                                & $C_\theta$        & FID       & $C_\theta$         & FID         & $C_\theta$        & FID       \\
Attacks                         &        43.05           &   5.4        &     48.79               &    2.67         &   40.33                &   6.61        \\
\rowcolor{gray!13} 
          & \multicolumn{6}{c}{\textbf{Black-box Attacks}} \\
{Crop} &    39.73               &   8.72        &      42.71              &   6.18          &      38.23             &      8.69   \\ 
{Blur} &    38.82               &   36.84        &    12.12                &   10.32           &      35.12             &   11.73        \\
{JPEG}     &   42.12                &  8.70         &      38.43             &     9.12        &     38.23              &   9.33        \\ 
{Noise}    &  40.26                 &  8.29         &         45.17           &     7.35        &    32.29                &    10.73      \\
Quantize                    &   37.17                &   11.60         &     43.27               &  5.61           &     39.72              &    8.71       \\ 

SR                  &    32.86              &   11.51   &  34.52   &      11.62             &     30.12        &       11.34                      \\
\rowcolor{gray!13} 
          & \multicolumn{6}{c}{\textbf{White-box Attacks}}  \\
Overwrite                     &     4.78             & 8.34           &   4.91                 &  8.83           &      4.73             &   9.71        \\ 
$\text{RPT}_{200}$                      &  4.91               &      5.47           &    4.52                &  3.52            &           4.59        &     6.65      \\ 
$\text{RPT}_{100}$                         &     4.44           &  5.56              &      4.21              &       3.90      &           4.47        &    6.75       \\ 
$\text{RPT}_{50}$                         &     4.38           &  8.07              &      4.38              &        15.32     &          4.16         &     14.47      \\\bottomrule
\end{tabular}
    \caption{The capacity and FID of all surveyed attacks. We ablate over multiple parameters for each attack and this table shows the points with the best (i.e., lowest) FID. $\text{RPT}_R$ stands for the Reverse Pivotal Tuning attack using $R$ real samples. }
    \label{tab:robustness-table}
\end{table}

\textbf{Ablation Study for RPT.} 
\Cref{fig:rpt-ablation} shows an ablation study over the amount of real, non-watermarked training data an attacker requires to remove a watermark. 
We measured these curves as follows: We randomly sample a set of $R$ real images and run the RPT attack encoded by \Cref{alg:rpt} with gradually increasing weight $\lambda_{\text{LPIPS}}$ on the LPIPS loss until the watermark is removed. 
Then, we compute the FID on $K=50,000$ images.
In all experiments, the watermark is eventually removed, but access to more data has a significant impact on the FID that is retained in the generator after the attack.
For StyleGAN2, we find that 80 images ($\approx 0.1\%$ of the training data) can remove the watermark at less than $0.3$ points of deterioration in FID, representing a visually imperceptible quality degradation. 
Our results demonstrate that an adaptive attacker with access to the generator's parameters can remove any watermark using only a small number of clean, non-watermarked images and can threaten the trustworthiness of watermarking. 
\section{Discussion}
This section discusses the limitations of watermarking and our study, the extension of our work to other image generators, and ethical considerations from releasing our attacks.

\textbf{Non-Cooperative Providers.}
Our study indicates that watermarking for image generators can be robust under restrictive threat models (see \Cref{sec:threat_model}), where the attacker is limited in their access to capable image generators.
For watermarking to effectively control misuse, every provider has to cooperate and watermark their generators before disclosing them.  
However, this is unlikely to occur in practice~\cite{westerlund2019emergence}.  
A capable adversary with access to sufficient training data and computational resources can always train their own generator without a watermark and provide it to others. 
It is important to acknowledge this inherent limitation of watermarking methods, as they cannot prevent this scenario.
Nonetheless, they can act as a deterrent to adversaries who cannot train their own generators from synthesizing harmful deepfakes.

\textbf{Watermarking from Scratch.} We focus on the robustness and undetectability of watermarking methods on pre-trained image generators due to the substantially higher scalability than watermarking from scratch.
Further research is needed to explore the potential impact of watermarking from scratch on robustness and undetectability. 

\textbf{Watermarking for Intellectual Property Protection}. Watermarking has also been used for Intellectual Protection (IP) protection of neural networks~\cite{ong2021protecting, lukasdeep, uchida2017embedding}. 
The IP protection threat model typically assumes white-box access of the adversary to the target generator and black-box API access of the defender to the adversary's generator to verify a watermark. 
For example, Ong et al.~\cite{ong2021protecting} embed a \emph{backdoor} into a generator that synthesizes images containing a watermark when the generator is queried with certain latent codes.
All watermarks evaluated in this paper assume no-box access to the target generator, meaning that our watermarks can be extracted from any of the generator's synthetic images. 
We show that existing no-box watermarks are not robust in the white-box setting, meaning that they are likely not suitable candidates for IP protection of generators in practice. 

\textbf{Other Generator Models}. 
Recently, image generators such as DALL$\cdot$E 2~\cite{ramesh2022hierarchical} based on the Transformer architecture~\cite{vaswani2017attention}, and latent diffusion models~\cite{rombach2022high} have been shown to also synthesize high-quality images. 
While OpenAI's DALL$\cdot$E 2 generator is only accessible through a black-box API\footnote{\url{https://openai.com/dall-e-2/}}, model checkpoints for latent diffusion are publicly available as a white-box\footnote{\url{https://github.com/CompVis/latent-diffusion}}. 
The utility of these checkpoints on FFHQ is comparable to that of the StyleGAN model checkpoints used in this paper (e.g., latent diffusion reports a FID score of $4.98$).
DALL$\cdot$E 2 and Latent Diffusion models also map from a latent space to images, but they accept auxiliary input such as text that controls the synthesis. 
While PTW is compatible with any image generator that maps latent codes to images, extending our work to other models requires developing a watermarking method to map to their parameters (see \Cref{sec:ptw-keygen}), which we leave to future work.
Our work demonstrates for one state-of-the-art generator model architecture (StyleGAN-based generators with a FID score of 2.1) that they can be watermarked effectively using our method. 

\textbf{Summary of Deepfake Detection.} Our research reveals that existing watermarks are not robust against an adversary with white-box access to the generator but can withstand a black-box adversary. 
This means that watermarking can be a viable solution for deterring deepfakes if the provider acts responsibly and the generator is provided through a black-box deployment. 
The provider's goal is to deter model misuse, which can be accomplished through several means. These include (1) monitoring and restricting queries, (2) relying on passive deepfake detection methods, or (3) implementing proactive methods such as watermarking. 
Monitoring lacks transparency and can deter usage of the model if the user does not trust the provider~\cite{tamkin2021understanding,ding2023towards}.
Passive detection methods may not detect deepfakes as the quality of synthesized images improves or the adversary adapts to existing detectors~\cite{dong2022think}. 
Active methods such as watermarking enable a different type of deployment that (1) does not require query monitoring and (2) remains applicable to future, higher-quality generators. 
The provider and the user agree on a mutually trusted third party to deploy the generator, who does not tamper with the watermark nor monitor the queries.
Our research suggests that such a black-box deployment represents a viable option in practice to prevent model misuse using existing watermarks.

\textbf{Ethical Consideration.} 
 Deep image generators can have potential negative societal impacts when misused, for instance, when generating harmful deepfakes.
Our contributions are intended to raise awareness about the limited trustworthiness of watermarking in potential future deployments of image generators rather than to undermine real systems.
While the attacks presented in our paper could be used to evade watermarking, thereby enabling misuse, we believe that sharing our attacks does not cause harm at this time since there are no known deployments of the presented watermarking methods. 
We aim to advance the development of watermarking methods that our attacks cannot break.

\section{Conclusion}
We propose Pivotal Tuning Watermarking (PTW), which is a scalable method for watermarking pre-trained image generators.
PTW is three orders of magnitude faster than related work and enables watermarking large generators efficiently without any training data.
We find that our watermark is not easily detectable without the secret watermarking key unless the attacker can access $\geq 100$ labeled non-watermarked and watermarked images. 
Our watermark shows robustness against attackers who are limited in their availability of similar-quality models. 
However, we are the first to show that watermarking for image generators cannot withstand white-box attackers.
Such an attacker can remove watermarks with almost no impact on image quality using less than $0.3\%$ of the training data with our adaptive Reverse Pivotal Tuning (RPT) attack. 
Our results challenge that watermarking prevents model misuse when the parameters of a generator are provided, showing that open-source watermarking in the \emph{no-box} setting remains a challenging problem.

\bibliographystyle{plain}
\bibliography{sample}

\appendix
\section{Attack Description}
\label{appendix:sec-attack-description}
This section describes the attacks and shows which parameters we explored in our grid search.
We refer to the image before attacking as the \emph{base} image $x$ and as the \emph{attacked} image after the attacker performs their attack $\Tilde{x}=\mathcal{A}(x, I)$ with auxiliary information $I$.
We find the range of evaluated parameters through experimentation by limiting the degradation of the attack on the visual quality of the images. 

\subsection{Black-box Attacks}
Black-box attacks assume only black-box API access to the \emph{target generator}, meaning that they can query the generator on arbitrary latent codes $z\in \mathcal{Z}$, but they have no knowledge of or control over the generator's parameters or its intermediate activations.

\begin{algorithm}
\caption{Super-Resolution Attack}
\begin{algorithmic}[1]
\Procedure{Super-Resolution}{$x, \rho$}
    \State $\Tilde{x} \gets \textsc{resize}(x, \lfloor \textsc{resolution}(x) \cdot \rho \rfloor)$
    \While{$\textsc{resolution}(\Tilde{x}) < \textsc{resolution}(x)$}
        \State $\Tilde{x} \gets \textsc{SR}(\Tilde{x})$ \Comment{\small apply SR model}
    \EndWhile
    \Return $\textsc{resize}(\Tilde{x}, \textsc{resolution}(x))$ 
\EndProcedure
\end{algorithmic}
\begin{algorithmic}[1]
\Procedure{Resolution}{$x$} 
    \Return resolution of x in pixels
\EndProcedure
\Procedure{Resize}{$x, d$} 
    \Return downsized image $x$ with resolution $d$
\EndProcedure
\end{algorithmic}
\label{alg:super-resolution}
\end{algorithm}
The black-box attacks can be described as follows.
\begin{itemize}
    \item \textbf{Crop}: First, the base image is center-cropped with a given cropping ratio $\rho\in (0,1]$ and then resizes the cropping back to the base image's original size. We experiment with cropping ratios $\rho\in [0.9,1]$. 
    \item \textbf{JPEG Compression}: This attack performs JPEG compression~\cite{wallace1992jpeg} on the base image with a quality $q$.
    A higher quality better preserves the visual quality of the image and we experiment with $q\in [80, 200]$.
    \item \textbf{Noise}: This attack adds Gaussian noise $\mathcal{N}(0, \sigma^2)$ to the image. 
    We experiment with $\sigma \in (0, 0.05]$.
    \item \textbf{Quantize}. This attack quantizes the number of states that a pixel can have. 
    We compute quantization by the following formula for a quantization $q\in [0,1]$.
    \begin{align}
        \textsc{Quantize}(x) = q \cdot \lfloor \nicefrac{x}{q} \rfloor
    \end{align}
    We experiment with quantization strengths $q\in [0.5, 1]$.
    \item \textbf{Super-Resolution}. Our Super-Resolution attack uses Latent Diffusion models~\cite{rombach2022high} provided through HuggingFace\footnote{\url{https://huggingface.co/CompVis/ldm-super-resolution-4x-openimages}}.
    The used model increases an image's resolution by a factor of $4$ through an optimization process. 
    We use Super-Resolution as a removal attack summarized by \Cref{alg:super-resolution}. 
    We experiment with scaling factors $\rho \in [0.125, 0.5]$. 
\end{itemize}
%
\subsection{White-box Attacks}
White-box attacks assume full access to the target generator's parameters, meaning that the adversary can issue virtually unlimited queries to the target generator and can control the generator's parameters and intermediate activations.

The white-box attacks can be described as follows.
\begin{itemize}
    \item \textbf{Overwriting}. The overwriting attack trains a decoder according to \Cref{alg:train-decoder} and overwrites the existing watermark using PTW described in \Cref{alg:ptw-embed}. 
    We experiment with different weights of the watermark embedding $\lambda_R$ for PTW (see \Cref{alg:ptw-embed}).
    Increasing the weight of the decoder's loss $\lambda_M$ results in a stronger perturbation of all images synthesized by the target generator. 
    We experiment with a weight $\lambda_M\in [0.5, 1.5]$. 
    \item \textbf{Reverse Pivotal Tuning} (RPT). 
    Our RPT attack is parameterized by the number of real, non-watermarked images $R$ available to the adversary. 
    We invert images in the generator's latent space by backpropagating the LPIPS loss between the currently generated image and the base image and updating the current latent code.  
    While other methods~\cite{zhu2020domain, roich2022pivotal} to invert real images can yield better results, backpropagation is simple and works well in practice. 
    During training, we iteratively synthesize images from a randomly sampled batch of inverted latent codes and optimize the LPIPS similarity between the generated and corresponding base images. 
    \Cref{alg:rpt} encodes our RPT attack. 
\end{itemize}

\section{Implementation Details}
\label{appendix:implementation}
This section describes the implementation details of our approach such as the hyperparameters we used to embed our watermarks or the reference to the (publicly available) pre-trained generator checkpoints. 
We make a fully working implementation of all methods surveyed in this paper available as open source.

\subsection{Generator Checkpoints}
We experiment with the following checkpoints. 
All checkpoints were made publicly available by the authors~\cite{karras2020analyzing, karras2021alias, sauer2022stylegan}. 
On \textbf{FFHQ-256}, we use the following models: StyleGAN2\footnote{\url{https://nvlabs-fi-cdn.nvidia.com/stylegan2-ada/pretrained/paper-fig7c-training-set-sweeps/ffhq70k-paper256-ada.pkl}}, StyleGAN-XL\footnote{\url{https://s3.eu-central-1.amazonaws.com/avg-projects/stylegan_xl/models/ffhq256.pkl}}, StyleGAN3\footnote{\url{https://api.ngc.nvidia.com/v2/models/nvidia/research/stylegan3/versions/1/files/stylegan3-t-ffhqu-256x256.pkl}}.
On \textbf{AFHQv2} we use the following models: StyleGAN2\footnote{\url{https://api.ngc.nvidia.com/v2/models/nvidia/research/stylegan2/versions/1/files/stylegan2-afhqv2-512x512.pkl}}, StyleGAN3\footnote{\url{https://api.ngc.nvidia.com/v2/models/nvidia/research/stylegan3/versions/1/files/stylegan3-t-afhqv2-512x512.pkl}}.
On \textbf{FFHQ-1024} we use the following models: StyleGAN2\footnote{\url{https://nvlabs-fi-cdn.nvidia.com/stylegan2-ada-pytorch/pretrained/ffhq.pkl}}, StyleGAN-XL\footnote{\url{https://s3.eu-central-1.amazonaws.com/avg-projects/stylegan_xl/models/ffhq1024.pkl}}, StyleGAN3\footnote{\url{https://api.ngc.nvidia.com/v2/models/nvidia/research/stylegan3/versions/1/files/stylegan3-t-ffhq-1024x1024.pkl}}

\subsection{Watermarking Parameters}
All watermarks in this paper are embedded with the following parameters.
We use an Adam optimizer~\cite{kingma2014adam} and we use a learning rate of $10^{-4}$ for the generator during PTW (see \Cref{alg:ptw-embed}).
We use the same generator learning rate for training a watermark decoder (see \Cref{alg:train-decoder}) and a learning rate of $10^{-3}$ for the watermark decoder. 
The watermark decoder training from \Cref{alg:train-decoder} contains a similarity loss and a binary cross-entropy loss for the watermark decoder. 
We scale the similarity loss with a weight $\lambda_{LPIPS}=1$ and the loss for the decoder with $\lambda_M=0.1$. 
We train with a batch size of 128 on FFHQ-256, a batch size of 64 on AFHQv2, and a batch size of 32 for FFHQ-1024.

\end{document}